\documentclass[11pt,a4paper]{article}
\usepackage[hyperref]{naaclhlt2019}
\usepackage{times}
\usepackage{latexsym}
\usepackage{amsmath,amssymb}
\usepackage[utf8]{inputenc} 

\usepackage{booktabs}
\usepackage{url}
\usepackage{graphicx}
\usepackage{csquotes}
\graphicspath{ {./graphics/} }

\usepackage{todonotes}

\aclfinalcopy % Uncomment this line for the final submission
\def\aclpaperid{1434} %  Enter the acl Paper ID here

\newcommand{\acl}{ACL-2018}
\newcommand{\ir}{iResp}
\newcommand{\dr}{dResp}
\newcommand{\kr}{kResp}
\newcommand{\os}{\texttt{OVAL}}
\newcommand{\border}{BRD}
\newcommand{\full}{Full}

\newcommand\blfootnote[1]{%
  \begingroup
  \renewcommand\thefootnote{}\footnote{#1}%
  \addtocounter{footnote}{-1}%
  \endgroup
}

\title{Does My Rebuttal Matter? Insights from a Major NLP Conference}

\author{Yang Gao$^*$, Steffen Eger$^*$, \\ 
\textbf{Ilia Kuznetsov, Iryna Gurevych} \\
Ubiquitous Knowledge Processing Lab \\ %(UKP-TUDA) \\
Department of Computer Science \\
Technische Universit{\" a}t Darmstadt \\
  {\tt www.ukp.tu-darmstadt.de} \\\And
  Yusuke Miyao\\
  Department of Computer Science \\
  Graduate School of Information \\ Science and Technology\\
  University of Tokyo \\
  {\tt yusuke@is.s.u-tokyo.ac.jp} \\}

\date{}

\begin{document}
\maketitle

\begin{abstract}
    Peer review is a core element of the scientific process, 
    particularly in conference-centered fields such as 
    ML and NLP.   
    %\emph{Natural Language Processing (NLP)}.
    However, only few studies have evaluated its properties empirically. % on a large scale. 
    Aiming to fill this gap, we present a %review 
    corpus that contains over 4k reviews and 1.2k author responses from %a major NLP conference, 
    \acl{}. 
    %the 56th  Annual  Meeting  of  the  Association  for Computational  Linguistics (\acl{}).
    We quantitatively and qualitatively assess the corpus. This includes a pilot study on paper weaknesses given by reviewers and on quality of author responses.
    We then focus on the role of the rebuttal phase, 
    and propose a novel %NLP 
    task to predict 
    %reviewers score-update directions (increase, decrease
    %or keep the initial score) from the peer reviews
    %and author responses.
    after-rebuttal (i.e., final) scores 
    from initial reviews and author responses. 
    %\SE{
    %Our results suggest that a reviewer's final score is largely determined by her initial score and the distance to the other reviewers' initial scores; yet, author responses do have a marginal 
    %%(though
    %(and 
    %statistically significant) influence on the final scores, especially for borderline papers. 
    %\SE{
    Although author responses do have a marginal (and statistically significant) influence on the final scores, especially for borderline papers, our results suggest that a reviewer's final score is largely determined by her initial score and the distance to the other reviewers' initial scores. 
    %}
    %We also \YG{discuss %the conformity 
    %some biases and weaknesses 
    %in the peer review process, which are %largely overlooked 
    %by previous research}. 
    %\SE{
    In this context, we discuss the \emph{conformity bias} inherent to peer reviewing, a bias that has largely been overlooked in previous research. %}
    We hope 
    %these findings and discussions 
    our analyses 
    will help better assess the usefulness of the %(expensive) 
    rebuttal phase in NLP conferences.
    %} 
    %Our results suggest that \YG{
    %the final scores are largely decided by
    %the peer review scores; 
    %author responses provide a marginal
    %(but statistically significant)
    %influence the reviewers' final scores. 
    %This finding sheds light on the usefulness
    %of the (expensive) rebuttal phase in 
    %NLP conferences.}
    %challengs the usefulness of the rebuttal
    %phase in NLP conferences.\todo{SE: hm, could mention that the rebuttal may matter}
    %
\end{abstract}

\section{Introduction}
Peer review is a widely adopted quality %assurance
control
%measure 
mechanism in which the value of scientific work is assessed by several reviewers with a similar level of competence. %It is opposed to an expert review in which the reviewer is more competent than the authors of the original work \todo{SE: any reference for this? Which reviewer is more competent than Dan Jurafsky?}, and has gained popularity due to its supposed higher efficiency and objectivity \todo{SE: same here?}.
\blfootnote{* Equal contribution.}
Although peer review has been at the core 
of the scientific process for at least 200 years \cite{Birukou:2011}, %comparatively little is known about it.
it is also a subject of debate: for instance,
%The few available studies 
it has been 
%\todo{SE: no, they just cite it, did not find it}
found that peer reviewing can hardly 
%some studies point out several problems of peer reviewing, such as its inability to 
recognize prospectively well-cited papers or 
%detect 
major flaws \cite{Ragone:2013}. %and %\cite{Ragone:2013},
Further, 
\citet{Langford:2015} observed substantial disagreement
between two sets of reviews on the same set of submissions
for the prestigious 
Conference on Neural Information Processing Systems
(NeurIPS) 2014.
%raising serious concerns on the validity of peer reviewing.
%and there recently emerges considerable debates in forums on 
%the validity and fairness of peer reviewing\footnote{
%\url{https://www.reddit.com/r/MachineLearning/comments/8ite3n/r_nips_2018_how_do_i_write_a_good_review/}, \url{https://www.reddit.com/r/MachineLearning/comments/924lkp/r_nips_2018_for_those_of_you_that_got_some_harsh/}, \url{https://twitter.com/aleks_madry/status/995358576268009473}}. 

%YG: come to rebuttal
The \emph{rebuttal} phase %is an important and
%widely used stage 
plays an important role 
in peer reviewing 
especially in top-tier conferences
in Natural Language Processing
(NLP). 
%and Machine Learning (ML). 
It allows authors to provide
\emph{responses} to address
the criticisms and questions raised in
the reviews %, so as to 
and to defend their work. 
%against unfair judgments.
%, so as to help the program committee
%members to better evaluate the quality of %submissions.
Although %statistics have shown that 
there is 
%anecdotal 
evidence that 
%a substantial proportion of 
reviewers do update their evaluations after the rebuttal phase\footnote{For example, see 
%at some conferences
%\footnote{For example, see %the statistics
%of before- and after-rebuttal scores for NAACL-2018
discussions 
at \url{https://naacl2018.wordpress.com/2018/02/04/analysis-of-long-paper-reviews/} and 
\url{https://acl2017.wordpress.com/2017/03/27/author-response-does-it-help/}.},
%\todo{YG: naacle blog comes here},
it remains unclear 
%whether the review update
%is largely due to the influence of the peer %reviews 
%%during the discussion after rebuttal 
%(i.e. the \emph{peer pressure}), or to the %author responses.
what causes them to do so, and especially, whether they react to the author responses per se, or rather adjust to the opinions of their co-reviewers (``\emph{peer pressure}'').

% This long-term and fundamental criticism has recently been supplemented by %the 
% studies in %specific, quickly developing 
% research areas such as machine learning (ML) and natural language processing (NLP), 
 %\todo{SE: not sure what extra information AI brings. If we're low on space, I suggest to drop it. IK: yes} 
% in which peer-reviewed conference publications have relatively high impact compared to 
% more established fields such as maths and physics.
 %\todo{SE: maybe it's not quite appropriate to call NLP/ML/AI less established. They probably only have other conventions than math, economics, etc.}.  
% The famous \emph{NIPS (Conference on Neural Information Processing Systems) 2014} experiment involved two organization committees, each with their own group of reviewers, for a subset of submitted papers. The substantial disagreement between the two committees concerning the accept/reject decisions \cite{Langford:2015} has raised concerns about the validity of peer review. Explosive growth of submission numbers to ML and NLP conferences poses an additional challenge: due to the increased workload and the lack of expert reviewers, conference organizers have to employ less experienced reviewers
% \footnote{See discussions at  \url{https://www.reddit.com/r/MachineLearning/comments/8ite3n/r_nips_2018_how_do_i_write_a_good_review/}, \url{https://www.reddit.com/r/MachineLearning/comments/924lkp/r_nips_2018_for_those_of_you_that_got_some_harsh/}, \url{https://twitter.com/aleks_madry/status/995358576268009473}}. % This further challenges  the quality of the review process.  
 %\todo{move footnote to refs?} %YG: I would go for sticking to footnote, as they are not proper publications; but indeed we need to adjust the alignment a bit, reduce the ugly big space

%v1.2, by YG
In order to %enhance the transparency and 
obtain further insights 
into the reviewing process,
especially regarding the role of the rebuttal phase in peer
reviewing, %in NLP and ML conferences,
in this work
we present and analyze a review corpus of %a major NLP conference: 
\emph{the 56th Annual Meeting of the Association for Computational Linguistics (\acl{})}.
%
%\YG{
Every reviewer/author was asked whether she 
consented to freely using her review/author-response 
for research purposes 
and publishing the data under an appropriate open-source license within 
at earliest 2 years from the acceptance deadline (see supplementary material
for the original consent agreement).
85\% reviewers and 31\% authors have consented to sharing their data.
%}
%Similar to the other major conferences in NLP, \acl{}  
%adopts the \emph{double-blind} reviewing model, i.e., neither the reviewers
%nor the authors know the identities of the other side, 
%and uses the rebuttal phase in its reviewing process.
The corpus comprises over 4k reviews (including review texts and scores) 
and 1.2k author responses. Uniquely, the corpus includes
both before- and after-rebuttal reviews for both accepted
and rejected papers, making it a highly valuable 
resource
for the community to study the role of the rebuttal phase. 
%
%According to the data-sharing agreement of \acl{}, 
%the corpus will become publicly available around July 2020,
%after proper anonymization. 
%\footnote{
The corpus as well as our source code and annotations
are publicly available at
%}
\url{https://github.com/UKPLab/naacl2019-does-my-rebuttal-matter}.
%}

%Our work has two main contributions. 
Our contributions are threefold.
First, in \S\ref{sec:corpus}, 
we %systemically 
assess the corpus both 
%\todo{SE: systemically or systematically? It's not super systematically so maybe systemically is even a better word?! :-)}
\emph{quantitatively}
(e.g., 
%studying the average length of ``good'' and ``bad''
%author responses, and studying the correlation between the 
correlating 
%\todo{SE: studying length of good author responses doesn't sound very scientific if it means to just state how long they are}
Overall Score with aspect scores such as Originality and Readability)
and \emph{qualitatively} (e.g., identifying key terms
that differentiate ``good'' from ``bad'' author responses,
annotating paper weaknesses given by reviewers,
and rating the quality %convincingness 
of individual author responses).
%
%The corpus will be publicly available
%at the earliest in approximately 2 years after \acl{}.
%checks, 
%at the earliest in 
%approximately 2 years after \acl{}.
%The size of the corpus and the multidisciplinary and 
%highly selective (acceptance rate 25\%)
%nature of the conference renders this corpus a valuable resource
%for NLP, AI and ML community.
%
%Second, we propose a novel NLP task called
%\emph{score update prediction}, namely, 
Second, in \S\ref{sec:score_classification}, we 
%quantitatively measure the
%influence of peer pressure and author responses
%on the score update, by developing 
develop 
a %simple 
model to predict whether a reviewer will 
%update her review scores
increase/decrease/keep her initial scores after the rebuttal. 
We do so in order to analyze and disentangle the sources of review updates during the rebuttal stage. 
%given the peer reviews and the author responses.
%The purpose of this task is to better understand the role
%of the rebuttal phase in the paper reviewing process.
%the decision-making process during the peer review.
%Besides the before-rebuttal reviews' scores, we also use 
%multiple text-based features in the prediction, 
%including specificity, politeness and convincingness of the
%author responses.
%We use two feature sets for this regression task: pre-rebuttal (i.e., initial) review scores and text-based features including specificity \cite{}, politeness \cite{politeness}, convincingness of the author responses as well as the %cosine similarity- IK: doesn't have to be cosine, it's just our modeling choice
%overlap between rebuttal and review text. 
%
%Interestingly, 
We find that %while 
factoring in the author responses 
only marginally %\SE{
(but statistically significantly) %} 
improves the classification performance,
%indeed brings an improvement
%in our classification task, %\todo{SE: todo, check this} 
and the score update decision is largely
determined by the %before-rebuttal 
scores of peer reviewers.
%This finding not only highlights the ``peer pressure'' effect
%in peer reviewing,
%\todo{SE: I think it doesn't highlight peer pressure unless we find that it's the mean that determines the scores} 
%but also challenges the %highly 
%
%We hope our findings will help the community
%re-evaluate the usefulness of the 
%(after all, costly)
%rebuttal phase. 
%
%\YG{%The above observations reflect certain 
Third, in \S\ref{sec:discussion}, we
discuss multiple types of biases in the score update process,
some of which potentially undermine the `crowd-wisdom' of
peer reviewing. %}
\label{sec:introduction}
\section{Related Work}
%Our work is based on two strands of research: meta science and analysis of peer reviews on the one hand, and opinion dynamics on the other hand. This section provides a short overview of the most relevant works in these fields.

%\subsection{Meta Science and Peer Reviews}

%Some ML conferences have started opening their
%reviews and author responses to the public.
%There exist multiple sources for obtaining reviews
%and author responses data.
Several sources provide review and author response data. 
Since 2013, the NeurIPS main conference publishes the
%from 
%in 2013 to publish 
%its
%accepted papers' 
reviews of accepted papers 
%(unclear whether they
%are before- or after-rebuttal reviews)
%\todo{SE: find out? YG: I checked their website, some review texts mentioned 'I have read the rebuttals', suggesting they are after-rebuttal. } 
and their author responses. 
However, these reviews only include
the review texts for after-rebuttal reviews.
Also, reviews of rejected papers and author responses
are not published.
Some Machine Learning and NLP conferences, for instance 
ICLR (International Conference on Learning Representations)
and ESWC (Extended Semantic Web Conference),
adopt the \emph{open review} model, which allows
anyone to access the reviews and author responses.
However, most major NLP conferences have not yet 
adopted the open-review model,
and the reviews and author responses
in open- and non-open-review venues
are likely to be different 
because people behave differently when their actions are observable \cite{andreoni2004public}. 
%, according to studies on 
%human behaviours in closed and open environment.
%\todo{YG: @SE, can you add a citation here?}

\citet{Kang:2018} provide a corpus of computer science 
papers from 
ACL, %(Annual Meeting of the Association for Computational Linguistics), 
NeurIPS, CoNLL (The SIGNLL Conference on Computational Natural Language Learning) and 
ICLR, together with the accept/reject decisions and reviews for a subset of the papers. 
They suggest several tasks with respective baselines, 
such as predicting review aspect scores from paper- 
and review-based features. 
However, their corpus contains neither before-rebuttal reviews
nor author responses, and the size
of their review set from NLP conferences 
(only 275 reviews from ACL-2017 and 39 reviews
from CoNLL-2016)  is much smaller than ours.

%\YG{
\citet{argument_review:2018} compile a corpus consisting of 14.2k reviews from
major NLP and machine learning conferences.
In addition, they annotate 10k argumentative propositions 
in 400 reviews, and train state-of-the-art proposition 
segmentation and classification models on the data.
But similar to \citet{Kang:2018}, their corpus does not include 
before-rebuttal reviews or author responses.
%}

%[One problem with their data is that their (high-quality) 
%ACL dataset is very small and biased due to self-selection. 
%We will discuss this below.] 
Several publications specifically address the peer reviewing process.
\citet{Falkenberg:2018} investigate what makes a paper 
review helpful to a journal editor within a specific scientific field. \citet{Birukou:2011} and \citet{Kovanis:2017} discuss 
the shortcomings of the review process in general, such 
as its inability to detect major flaws in papers \cite{Godlee:1998} 
and its ineffectiveness in selecting papers that will have 
high citation counts in the future \cite{Ragone:2013}. 
They discuss alternatives to the standard review 
process such as crowd-based reviewing and
review-sharing, i.e., resubmitting a rejected work to another venue
along with its past reviews.
%for journal papers, review-sharing in which 
%rejected manuscripts are resubmitted along with their past 
%reviews to other journals. 
\citet{Ragone:2013} analyze peer reviews across 
nine anonymized computer science conferences and, 
among others, identify reviewer biases of multiple types
(affiliation, gender, geographical, as well as rating 
bias: consistently giving higher or lower scores 
than other reviewers) and propose means for debiasing reviews.
However, none of these works quantitatively measures the
influence of the rebuttal phase on the final review scores,
nor do they provide any corpora facilitating such studies.

%Thus, existing peer review studies focus either on journal submissions, for which the peer review process is typically considerably different from that of conferences, or on specific aspects of the review process such as reviewer bias. To our knowledge, we are the first to study the problem of estimating the post-rebuttal score from the pre-rebuttal score and we are thereby the first to empirically assess the value of the rebuttal in the review process.\todo{SE: this paragraph must be sharpened}

%Our work connects to two strands of research, 1) opinion dynamics, that is, %the 
%individuals' change of opinions based on exchanges within their `social network', and 2) meta science and the analysis of peer reviews. We discuss these in the following. 

Our work is also related to \emph{meta science},
which studies the scientific process in general, i.e., how scientific information is created, verified and distributed (cf. \citet{Fortunato:2018}). 
In this context, our work can be seen 
%investigation into 
%the role of the rebuttal phase in peer 
%reviewing can be seen
as a study on how scientific information is verified.
%verification.
%meta science regarding how scientific information
%is verified.
%For example, \citet{Fortunato:2018} review general aspects of the `science-of-science' such as a tendency towards multi-authored papers over the last decades, citation distributions of papers, and individual researchers' career dynamics. 

\label{sec:related}
\section{Review Corpus}
\label{sec:corpus}
\acl{} adopts a reviewing workflow similar to 
that of other major NLP conferences:
after paper assignment, typically three reviewers
evaluate a paper independently. 
After the rebuttal, reviewers can access the author
responses and other peer
reviews, and discuss their viewpoints. 
%to reach an agreement.
Reviews include both \emph{scores}
(Overall Score \os{}, 
Reviewer Confidence \texttt{CONF},
Soundness \texttt{SND}, 
Substance \texttt{SBS}, Originality
\texttt{ORG}, Meaningful Comparison \texttt{CMP} 
and Readability \texttt{RDB}) and free-text comments.
\os{} are integers in $[1,6]$,
while all other scores are integers in $[1,5]$.
%Note that all scores are in the range of $[1,6]$.
%\todo{SE: only the overall score, as far as I remember. Also: this reveals us uniquely. If we don't need this information, drop it.
%YG: double checked, you are right regarding the scores. I think the score ranges are important, 
%because we define the borderlin as those between
%3 and 4.5; in the system of [1,5], 4.5
%is too high for borderline.}

%
%We first briefly recap the review workflow in 
%\acl{}
%the conference 
%in \S\ref{subsec:workflow},
We first provide an overview of our corpus in \S\ref{subsec:corpus_overview}, and
then present analyses for the 
reviews and author responses
in %\S\ref{subsec:submissions}, 
\S\ref{subsec:reviews_analyses} and \S\ref{subsec:rebuttal_analyses}, respectively.

\subsection{Overview of the Corpus}
\label{subsec:corpus_overview}
%To construct our corpus, we have collaborated with the 
%chairs of \acl{} to obtain the
%Our corpus consists of the 
%\todo{YG: change based on IG's comments}
%reviews and author responses opted in to the
%data collection program, % in \acl{},
%accounting for around 80\% of all
%submitted reviews and responses.
%opted-in reviews and author responses from \acl{}. 
%Around 80\%\footnote{Exact percentages are hidden for anonymity reasons, and will be disclosed upon acceptance.} 
%of the reviewers and authors in \acl{} opted in the data 
%collection program.
The corpus has three parts: the before-rebuttal
reviews (including review texts and 
scores), the after-rebuttal reviews, and 
the author responses.
%Note that 
The corpus does not contain the submissions,
nor the information of the reviewers,
e.g., their gender, country, affiliation or seniority level; 
nevertheless, 
%we analyze the high-frequent
%n-grams in accepted papers and report the 
%results in the supplementary material.
we perform some analyses on the submissions and
the reviewers' information
and present the statistics in the supplementary material.

%YG: new added about the opt in/out process
%Before reviewers and authors input their reviews and responses, respectively, 
%they can choose whether to opt in to the data collection and sharing program or not. 
%1275 out of 1610 (79.2\%) reviewers 
%and 1198 out of 1551 (77.2\%) authors opted in the program. 
%
%\todo{SE: maybe we should talk about the opt-in possibility}

%As for reviews, our corpus includes 3875 before-rebuttal reviews
%from 1213 reviewers and 4054 after-rebuttal reviews from 1275 reviewers.
%The size of the after-rebuttal reviews is larger because 
%it also includes the
%additional reviews submitted during the discussion phase and
%the late reviews submitted after the rebuttal phase started.

\begin{figure}
\includegraphics[scale=0.4]{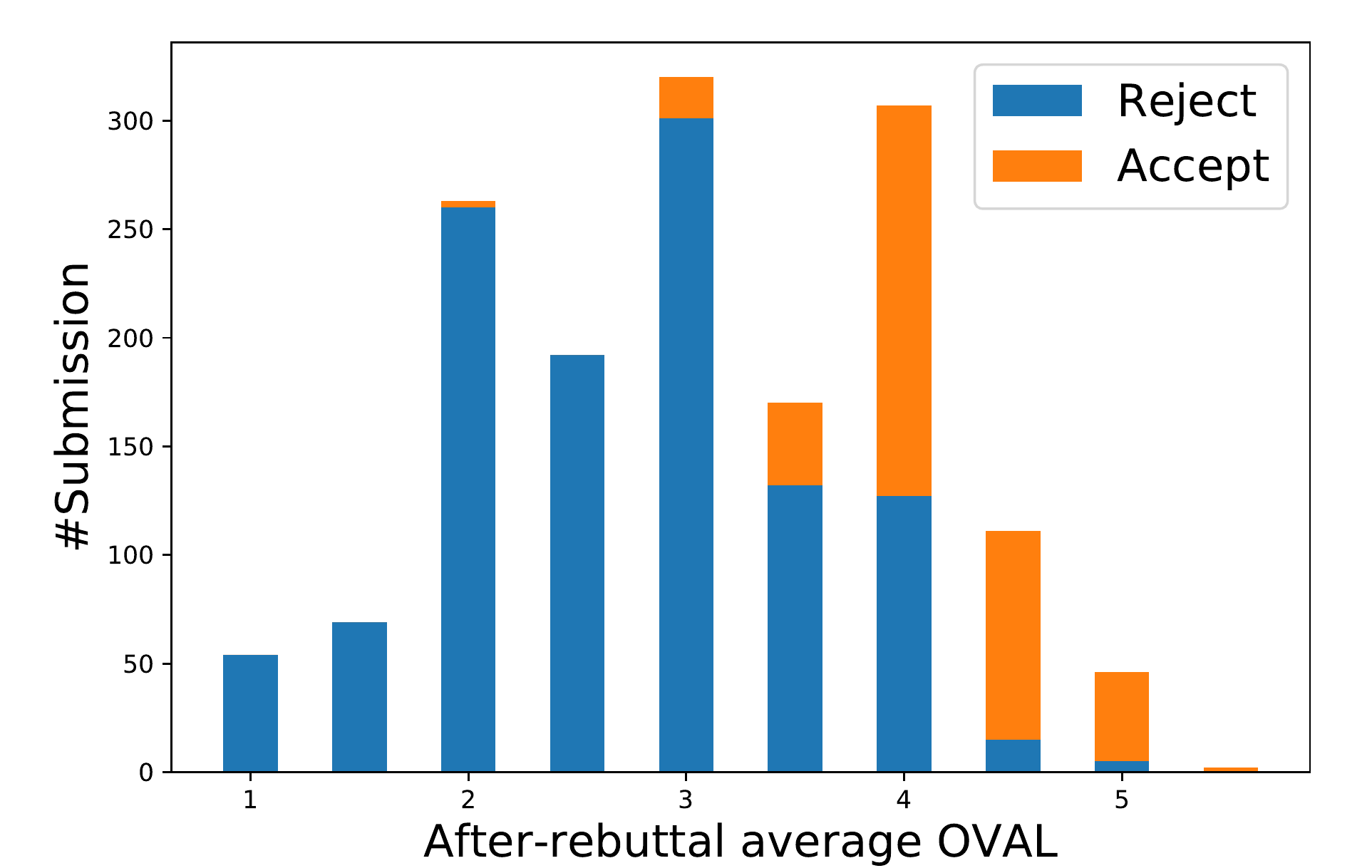}
\caption{Distribution of accept/reject decisions.
%among the 1542 submissions covered in our corpus.}
}
\label{fig:acc_rej_hist}
\end{figure}

Basic statistics of our corpus are summarized in 
Table \ref{tab:corpus_overview}.
1542 submissions (1016 long, 526 short) have at least 
one review opted in. 1538 submissions have
at least one before- and one after-rebuttal review opted in.
Among the 1542 submissions, 380 submissions 
(24.6\%) were accepted: 255 long, 125 short, 
and the remaining 1162 were rejected: 761 long, 401 short. 
%For the reviewers who opted in, their messages sent to
%other reviewers during the discussion phase are also included in our corpus.
The distribution of their accept/reject decisions
is illustrated in Fig. \ref{fig:acc_rej_hist}.

%As for author responses, 1198 %submissions' 
%rebuttals are opted-in;
%among them, 2 submissions have no reviews opted in and 1 submission 
%has empty author responses; these 3 submissions are hence removed from our corpus.\todo{SE: can think about making this a table without explanations.}

\begin{table}[h]
    \small
    \centering
    \begin{tabular}{p{2.2cm}|p{3.6cm}}
    \toprule
     Category & Size \\
     \midrule
     Before-rebuttal reviews & 3875 (1213 reviewers, 1538 submissions) \\
     \hline
     After-rebuttal reviews & 4054 (1275 reviewers, 1542 submissions) \\
     \hline
     Author responses & 1227 (499 submissions) \\
     \bottomrule
    \end{tabular}
     \caption{Statistics of the \acl{} corpus. 
     Some reviewers submitted their reviews after
     the rebuttal started, hence the size of 
     the after-rebuttal reviews is larger than that
     of the before-rebuttal reviews.}
    \label{tab:corpus_overview}
\end{table}

\subsection{Reviews}
\label{subsec:reviews_analyses}
%Conference reviewers are required to rate particular aspects of the papers under review, as well as to assign the final paper score and state their confidence level. The contribution of the individual aspect scores to the final score varies among reviewers.
%We present some statistics of the reviews 
%based on both the review texts and the review %scores.

\begin{figure}
\includegraphics[scale=0.55]{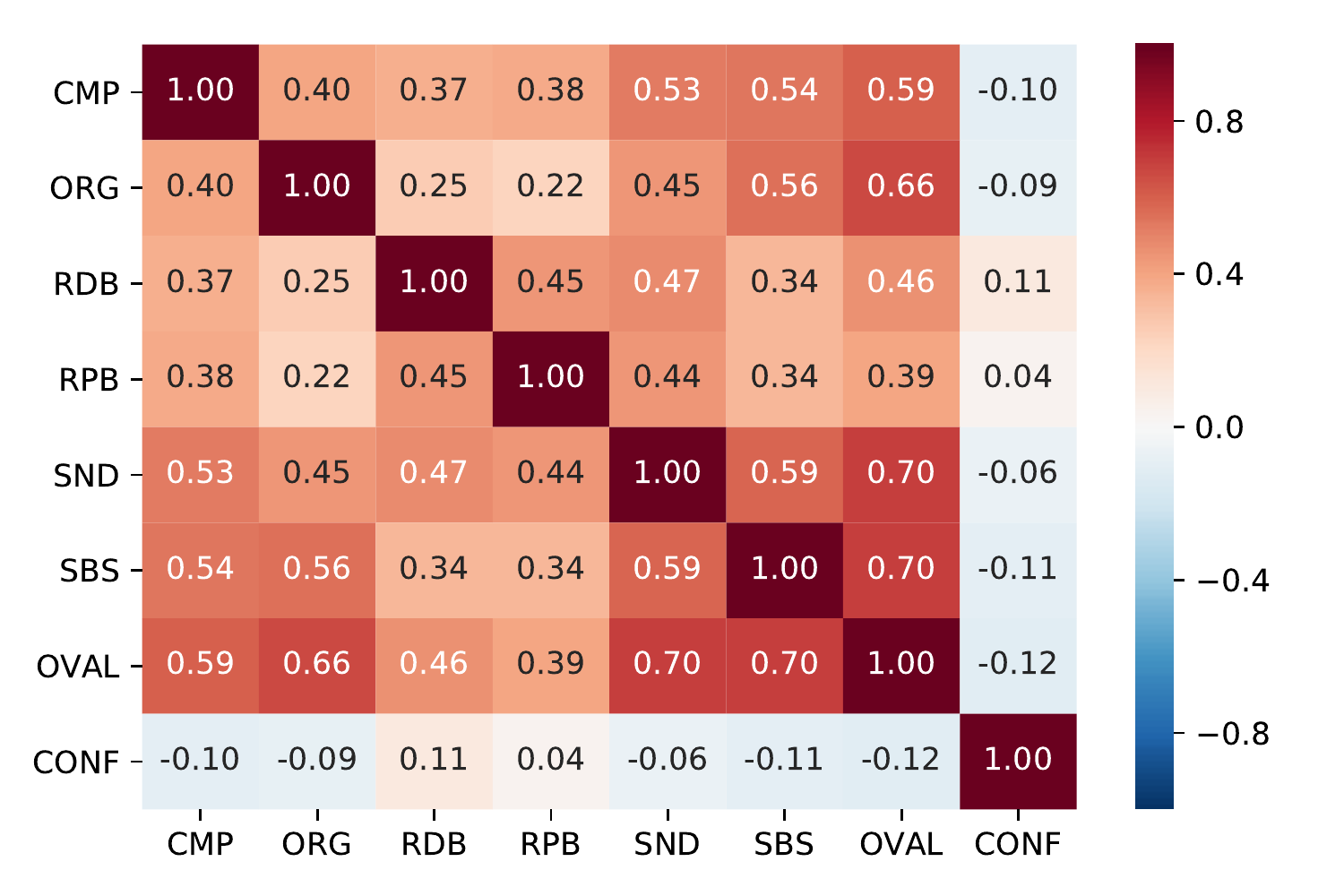}
\caption{Score correlation matrix.}
\label{fig:corr_acl_18}
\end{figure}

%In line with \newcite{Kang:2018}, we use 
%Pearson correlation %coefficient 
\paragraph{Score Correlation.}
In line with \newcite{Kang:2018}, we first assess the 
impact of 
individual aspect scores on the overall score
by measuring their Pearson correlation,
%Correlations between the final scores and the aspect scores are 
illustrated in Fig. \ref{fig:corr_acl_18}.
%
%We replicate and extend their analysis by providing full correlation matrices for both ACL-2017 train (Fig. \ref{fig:corr_peerread}) the ACL-2018 review data (Fig.  \ref{fig:corr_acl_18}). 
We find that \texttt{OVAL} is most strongly correlated 
with \texttt{SND} and \texttt{SBS}, 
followed by \texttt{ORG}
and \texttt{CMP}. %Meaningful Comparison. 
\texttt{CONF} shows weak positive 
correlation to \texttt{RDB}: %Readability: 
the less readable a paper 
is, the less confident the reviewers will be. 
Note that our correlation results are different from 
those reported by \newcite{Kang:2018}, 
who report that the \texttt{OVAL} has low Pearson correlation
with \texttt{SND} (0.01) and \texttt{ORG} (0.08).
While the differences might be caused by a  variation in aspect definitions, 
%attribute it mostly to the order-or-magnitude size difference between the datasets.
we believe that our estimate is more reliable as 
the dataset analyzed in \citet{Kang:2018} is substantially smaller than ours.
%YG: do not directly compare to ACL, which discloses our identity
%our dataset is substantially larger than the ACL section of \citet{Kang:2018}. 

%Note that \newcite{Kang:2018} also performed 
%a similar correlation analysis
%on a small subset of the ACL 2017 data. They also find that
%the Overall Score has a strong correlation with \texttt{SBS} (0.59), but that its 
%correlation to \texttt{SND} and \texttt{ORG} is quite weak: 
%0.01 and 0.08, respectively. 
%This appears peculiar since originality appears intuitively very 
%important 
%for an ML or NLP paper. We hypothesize 
%this result is an artefact of their 
%small corpus size, %of the data they use,
%which only includes 275 reviews for 137 submissions.
%While the differences might be caused by the variation in aspect definitions, we attribute it mostly to the order-or-magnitude size difference between the datasets.

\paragraph{Review Texts.}
%As for the review texts, 
\acl{} adopts the novel \emph{argument-based}
review template, which asks reviewers to provide 
\emph{positive} and \emph{negative} arguments
for and against the submission, respectively. 
%\todo{SE: actually, that reveals us again :D. YG: NAACL-18 also  uses the template} 
In addition, reviewers can also list their questions to
the authors in the \emph{questions} section
of the review template.
Most reviewers made good use of the argument-based
template: among the 4054 after-rebuttal reviews, 3258 (80.4\%) provide 
positive arguments, 3344 (82.5\%) provide negative arguments, 
and 1627 (40.1\%) provide questions.
The number and length of arguments/questions are summarized
in Table \ref{tab:arg_num}.

\begin{table}[h]
    \small
    \centering
    \begin{tabular}{c c c}
        \toprule
         Component & Number & Length (token) \\
         \midrule
         Pos. Arg. & 1.92$\pm$1.31 &  22$\pm$17 \\ 
         Neg. Arg. & 2.38$\pm$1.56 & 56$\pm$53 \\ 
         Questions & 0.87$\pm$1.36 & 35$\pm$31 \\
         \bottomrule
    \end{tabular}
    \caption{Numbers and lengths of different components in each review
    (mean$\pm$standard deviation).}
    \label{tab:arg_num}
\end{table}

\paragraph{Score Changes.}
Table \ref{tab:score_update} shows how many reviews
increase (INC), decrease (DEC) or keep (KEEP) their
overall scores after rebuttal.
For the 227 papers that receive
at least one INC review (first row in Table \ref{tab:score_update}),
their acceptance rate is
49.8\%, much higher than those 221 papers with at least one
DEC (7.2\%) and those 1119 papers 
with no score update (22.8\%).
Hence, the score update
has a large impact on the final accept/reject decision.
Note that 29 papers receive both INC and DEC reviews,
of which five were accepted finally.

\begin{table}[]
    \small
    \centering
    \begin{tabular}{l | c c c c c}
        \toprule
         Type & Num. & \#Paper & Acpt.\% &  $\Delta_{\texttt{OVAL}}$ \\
         \midrule
         INC & 245 & 227 & 49.8 & 2.65 $\rightarrow$ 3.76 \\
         DEC & 248 & 221 & 7.2  & 4.17 $\rightarrow$ 3.04 \\
         KEEP & 3377 & 1119 & 22.8 & 3.13 $\rightarrow$ 3.13  \\
         \midrule 
         Total & 3870 & 1538 & 24.7 & 3.17 $\rightarrow$ 3.17 \\
        \bottomrule
    \end{tabular}
    \caption{Statistics of different types of reviews.}
    \label{tab:score_update}
\end{table}

\begin{figure}
\includegraphics[scale=0.5]{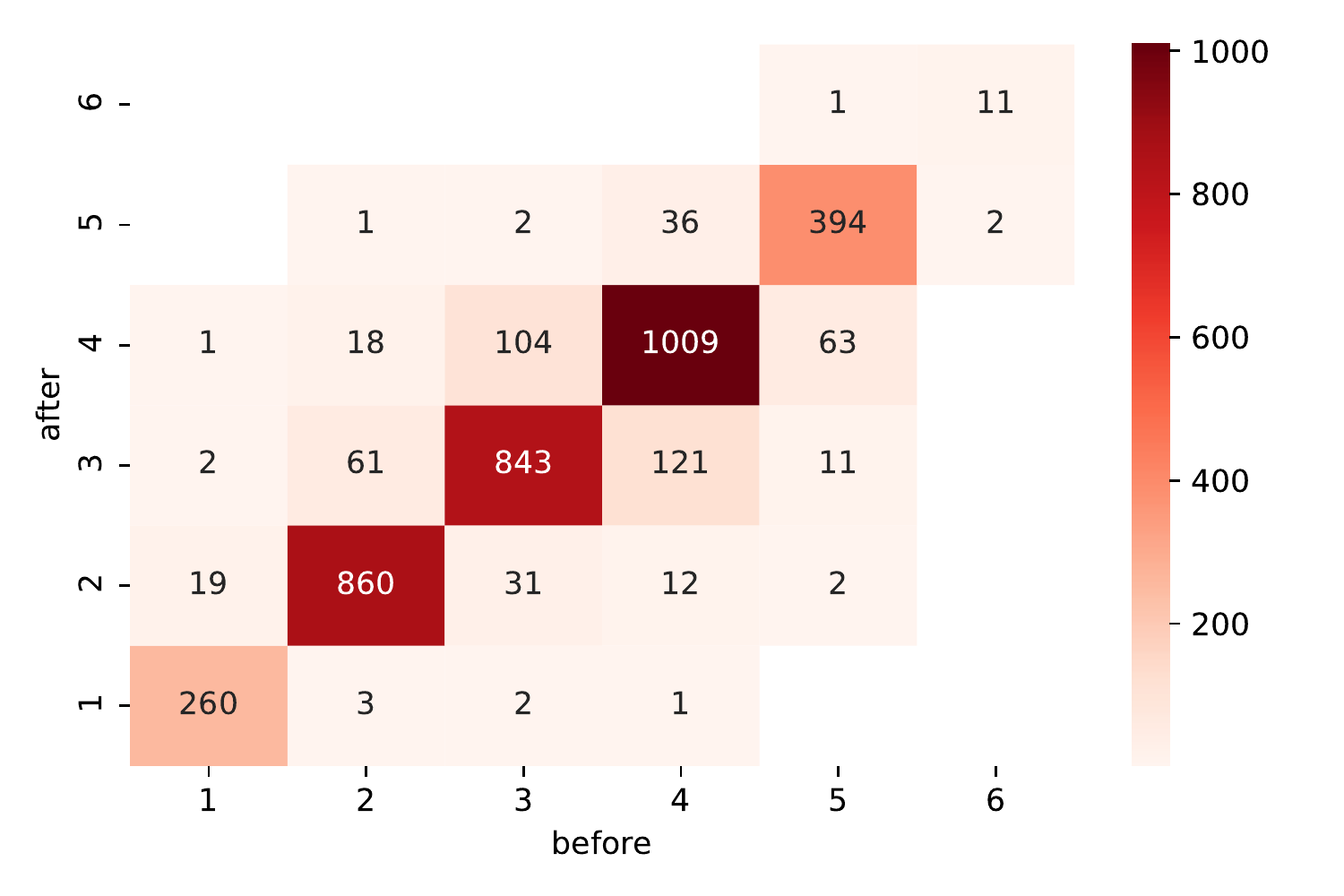}
\caption{Before vs after rebuttal \os{}.}
\label{fig:self_before_after}
\end{figure}

Fig.~\ref{fig:self_before_after} summarizes the 
\os{} updates.
Most reviewers stick to their initial scores
after rebuttal. 
For those who update, the score change usually amounts to just one point in absolute value.
However, most updates happen in the borderline 
area (overall score 3-4) where the score update 
might influence the overall acceptance decision.
We find that the changes in aspect 
scores occur much less often than
the changes in overall scores: 
only 5\% of the reviews have any of the aspect 
scores updated after rebuttal, and only 1\% of the 
reviews change the confidence value. In these rare cases, aspect score changes are consistent with
their \os{} changes, e.g.,  
if the \os{} increases, no aspect score decreases.

%On average, each review contains 1.92$\pm$1.31 
%(standard deviation) positive arguments,  
%2.38$\pm$1.56 negative arguments and 0.87$\pm$1.36 questions.
%Each positive argument on average contains 22$\pm$17 
%tokens,  each negative argument contains 56$\pm$53 tokens,
%and each question contains 35$\pm$31 tokens. 
%PCs and ACs have found the arguments in the reviews 
%highly helpful in making the final recommendations.
%(cite ACL18 
%proceedings, todo).\todo{SE: omit?}

\iffalse
\begin{figure}
\includegraphics[scale=0.6]{corr_peerread_acl17_train.png}
\caption{Aspect correlation PeerRead (ACL 2017 train)}
\label{fig:corr_peerread}
\end{figure}
\fi

\begin{figure}
\includegraphics[width=0.5\textwidth]{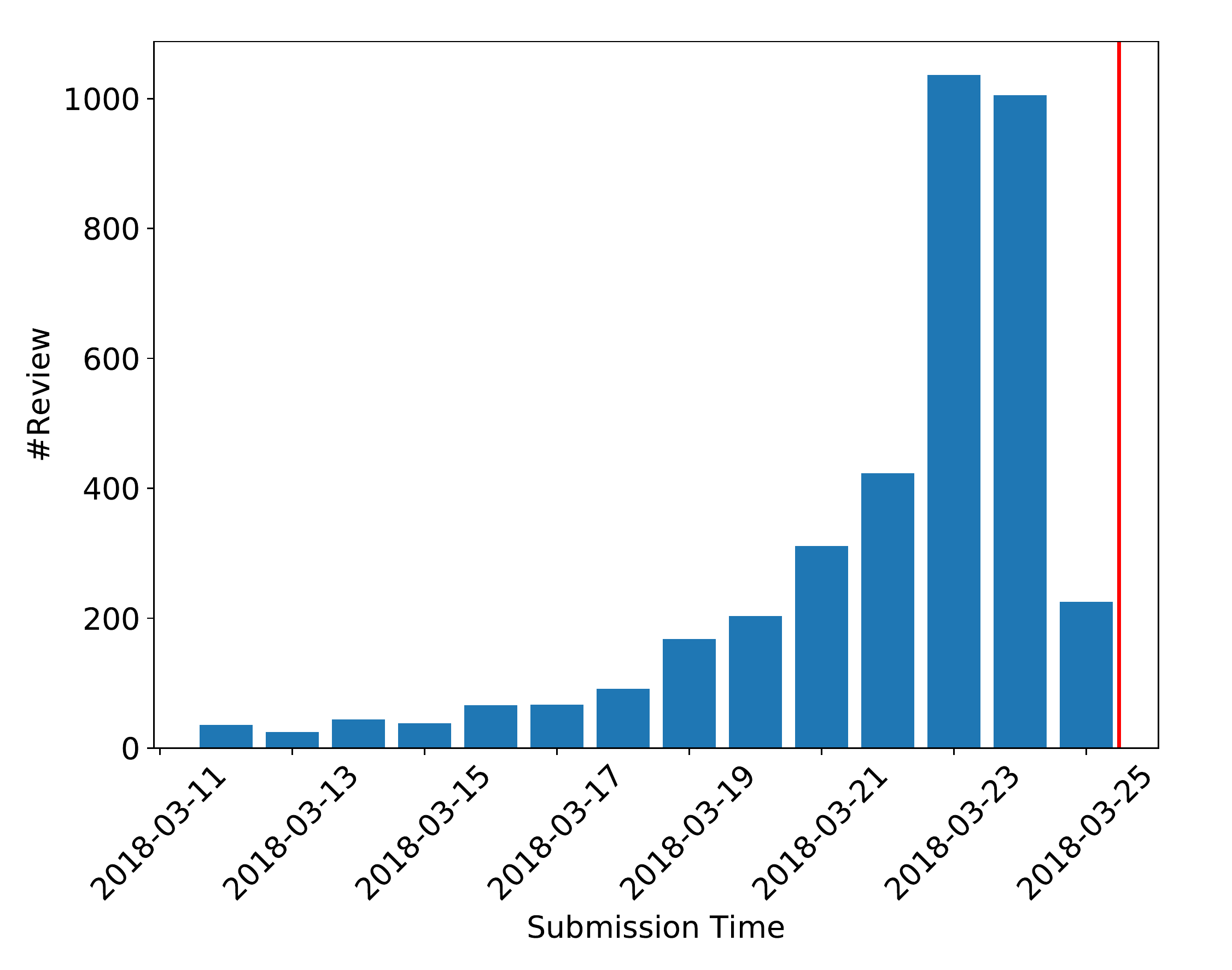}
\caption{Distribution of review submission time. The review submission 
deadline (26th March 2018) is marked as the red vertical line
towards the right end.}
\label{fig:review_time}
\end{figure}

\paragraph{Submission Time.}
%\todo{SE: Interesting. Anything that this might tell us? Why do do good guys submit early? YG: add a speculation, please see whether
%it makes sense}
Fig.~\ref{fig:review_time}  illustrates the distribution of the first submission time of reviews. 
51.6\% reviews were submitted
within the last 3 days before the deadline.
We also find that the mean submission time of the INC reviews
is around 20 hours earlier than that of the 
DEC reviews, and the difference is statistically
significant (p-value 0.009, double-tailed t-test).
%
%We speculate that this is because reviewers tend to be
%more selective at the beginning and gradually 
%calibrate their selection standard afterwards.
%Hence, for the papers that are reviewed at an early time
%point, reviewers tend to give them lower initial scores,
%but are also more likely to calibrate their scores
%after they read the peer reviews
%and author responses.
%
%However, there is no significant difference
%between the mean submission time between
%INC and KEEP (p-value 0.040) or DEC and KEEP %(p-value 0.185).
Moreover, we find that submission time is weakly positively correlated with initial score, which means that reviewers who submit early have slightly lower scores on average, which may explain their tendency to increase their scores later on, given our results in \S\ref{sec:score_classification}. A reason why early submitters have lower scores may be that it takes less time to reject a paper, as the majority of papers is rejected anyway. 

%
%\YG{Taking these observations together with
%the fact that the mean initial scores of INC %reviews (2.65)
%are significantly lower than that of DEC reviews %(4.17),
%we speculate that ... } \todo{YG: not sure we have %the space 
%for speculations here. if we add this, which else %should be removed ...}

\paragraph{Criticism in Reviews.}

\begin{table}
\small
\centering
%\begin{tabular}{r c c c c c c}
%    \toprule
%     Eval & Writing & Data & Nov & Comp \\
%      29\% & 12\% & 11\% & 10\% & 10\% \\ 
%      \bottomrule
%\end{tabular}
\begin{tabular}{r c c c c c c}
    \toprule
     Eval & Writing & Nov & Data & Motivation \\
      28\% & 18\% & 8\% & 8\% & 5\% \\ 
      \bottomrule
\end{tabular}
\caption{Frequent weakness types identified in reviews.}
\label{tab:weakness_dist}
\end{table}

To study the most common paper weaknesses
identified in reviews, 
%What are the most common weaknesses of papers pointed out by the reviewers? 
%To gain qualitative insights, 
we manually assess about 300 weakness statements from the reviews. Table \ref{tab:weakness_dist} summarizes the main results, excluding concerns about \emph{Technical weaknesses}. 
In our sample, most weaknesses %(29\%)
refer to  \emph{Evaluation \& Analysis}, i.e.,  criticize  the lack of: error analysis, ablation tests, significance tests, 
human evaluations (opposed to indirect measures such as BLEU) and strong baselines as well as insufficient comparisons (either external or internal). 
%lack of motivation. 
%The second 
Other frequent targets of criticism are \emph{Writing} quality, %(17~cases)
as well as \emph{Data}: % (16 cases). 
e.g., too few datasets being used (only English data
or only synthetic data), missing agreement scores for newly labeled datasets, and
resources not being publicly available. 
%Papers also get criticized for 
Reviewers also criticize the lack of 
\emph{Novelty} and proper \emph{Motivation} of approaches. % (14)
%and insufficient \emph{Comparisons} %(mostly to other work, but also faulty internal comparison of models)
% (either external or internal).% (14). 
%\textbf{Writing} (11) and \textbf{Readability} (6) 
%are also frequently mentioned (inconsistent formulations; ``hard to read''; weak/wrong explanations). 

\subsection{Author Responses}
\label{subsec:rebuttal_analyses}

\begin{table}[]
    \small
    \centering
    \begin{tabular}{c c c}
        \toprule
         Type & Num. & Length (token)  \\
         \midrule
         \ir & 100 & 373$\pm$191 \\
         \dr & 80 & 260 $\pm$140 \\ 
         \kr & 1047 & 297$\pm$182 \\
         \midrule
         Total & 1227 & 300$\pm$181 \\
         \bottomrule
    \end{tabular}
    \caption{Statistics of author responses
    (mean$\pm$standard deviation for Length).}
    \label{tab:responses}
\end{table}

%If a reviewer increases or decreases her score after the rebuttal,
%we will refer to the corresponding response
%as \emph{good} or \emph{bad} response, respectively.
%Rebuttals that do not result in either score increase or decrease
%are termed as \emph{normal} responses.
%\todo[inline]{SE: which of the two definitions is it?
%YG: the fine-grained one}
%We find good responses are usually longer than 
%normal and bad responses: the average token number in 
%a good, bad and normal response is 382$\pm$192, %.18 (stdev 192.66), 
%%268.29 (139.70) 
%268$\pm$139 and 
%%303.99 (182.08),
%303$\pm$182,  respectively.

%499 submissions opted in their author responses.
We align author responses with their corresponding 
reviews (if opted in), and term the author responses 
corresponding to INC, DEC and KEEP reviews as \ir{}, \dr{} and \kr{},
respectively.
Table \ref{tab:responses} presents an overview on these groups.
%Note that the word limit is 1000 for the
%author response of each submission,
%regardless how many reviews the submission
%receives.

%respectively. 
To qualitatively compare \ir{}s and \dr{}s, we extract and rank n-grams in both \ir{}s and \dr{}s 
according to the \emph{log-likelihood ratio} 
(LLR) statistic \cite{Dunning:1993}, 
treating \ir{}s and \dr{}s as two corpora\footnote{We only include n-grams that appear in at least
7 different author responses.}. 
The results are reported in Table~\ref{table:ngrams_in_rebuttal}.
We find both \ir{}s and \dr{}s express gratitude 
and promise revisions in the final versions,
but \ir{}s address review questions and
criticisms by referring back to certain lines and tables in the original paper
%(in line DIGIT, in table DIGIT) 
while \dr{}s fail to do so. 
We revisit these differences 
%between \ir{}s and \dr{}s  
in \S\ref{sec:scorechange}.
%by comparing multiple facets of them, 
%including their specificity, politeness and convincingness.\todo{SE: can also refer back here to previous study}

\iffalse
\begin{table}
\centering
\begin{tabular}{c|c|c}
     rev & before & after \\
     \hline
     1 & 128 & 85 \\
     2 & 512 & 468 \\
     3 & 870 & 928 \\
     4 & 27 & 56 \\
     5 & 1 & 5 \\
\end{tabular}
\caption{Reviews per paper before and after rebuttal}
\label{table:reviews_per_paper}
\end{table}
\fi

\begin{table}[!htb]
    \centering
    \footnotesize
    \begin{tabular}{l|l}
         \toprule
         \ir{}s  &  \dr{}s \\ \midrule
         the final version & thanks for your \\
         in line DIGIT & DIGIT reply to \\
         in table DIGIT & to question DIGIT \\
         in the final & will add more \\
         for example the & DIGIT we will \\
         in order to & thank the reviewer \\
         final version EOS & argument DIGIT reply \\
         due to space & due to the \\
         for your comments & paper is accepted \\
         camera ready version & the revised version \\
         DIGIT and DIGIT & we agree that \\
         \bottomrule
    \end{tabular}
    \caption{Top trigrams based on LLR ranking. All digits
    were replaced by DIGIT. EOS: end of sentence.}
    \label{table:ngrams_in_rebuttal}
\end{table}

%\todo{IK: hm here the transition could have been smoother. Is it the same convincingness we tackle later, or a different one? YG: different, here is the cvc for the whole response
%with consideration of the reviews,
%later the machine generated score is on sentence level.}
To gain further insights, we analyze 
the \emph{quality} of the author responses 
to the 300 weakness statements from Table \ref{tab:weakness_dist}.
%\todo{YG: I change it to quality to avoid confusion}
We advertised no formal definition of quality, 
and assessed a subjective, perceived 
quality score in a range from 1 (low) 
to 10 (high). We find that 
the \emph{weak} author responses (scores 1-3) 
are substantially shorter than the \emph{strong} ones
(scores 8-10): the average token
number in weak and strong responses
% 44 and 102
are 53 and 90, respectively.
%low score responses have 44 
%tokens on average and high score responses have 102.
%tokens on average. 
Responses evaluated as weak are 
%substantially 
less specific and make vague promises 
(\textit{``Thanks for the suggestion, we will try this in the 
camera-ready''}), off-topic (addressing different 
points than those raised by the reviewer), or apologetic
%find excuses 
%argue that an important issue could not be done 
%due to space or time limitations 
(\textit{``the deadline was very close'', ``our native language is not English''}). Interestingly, 
with some exceptions (\textit{``We take your review as an example of bad writing''}), 
the weak responses %labeled by us as weak 
are usually polite and admit 
%a fault, 
the weaknesses suggested by the reviewers,
but they tend not to detail how they would address the weaknesses.
%did not go into details how to address this fault 
%or only made vague promises to revise in the camera-ready version.
Strong responses, in contrast, are specific (referring to specific line numbers in the submission, as well as providing numerical values), detailed, 
longer, and often do not agree with the criticism, 
but explain why the reviewer's requirement 
is hard to meet or beyond the scope of the work.
%\todo{SE: those responses that we scored highly/lowly, did they get increases/decreases? Is there a correlation?}

%\begin{figure}
%\includegraphics[scale=0.6]{graphics/own_vs_others_decision.png}
%\caption{Another plot, color is decision (raise, drop, keep), X is the distance to other %reviewers average, Y is probability within a class. Let's discuss this I'm not sure about %that one. IK}
%\label{fig:self_others_binary}
%\end{figure}
\label{sec:quantitative}
\section{After-Rebuttal Score Prediction}
\label{sec:score_classification}
To measure the influence of different factors
on the score update decisions, %in this section
we propose and study the \emph{after-rebuttal score 
prediction} task. 
Because most score updates after rebuttal do not exceed 1 point
(see Fig. \ref{fig:self_before_after}), we formulate this problem
as a classification task.
Specifically, given a before-rebuttal review, 
%its peer reviews \todo{SE: peer reviews of the before-rebuttal review?}
its corresponding author response and 
other peer reviews, we try to
predict whether the reviewer will increase (INC),
decrease (DEC) or keep (KEEP) her
overall score after the rebuttal.
We avoid predicting the final accept/reject decisions
because they are not only based on the final scores
(see Fig.~\ref{fig:acc_rej_hist},
where a few low-score papers are accepted while some
high-score papers are rejected),
but also based on additional factors such as the 
balance of areas and diversity of papers,
which are difficult to measure.
The score updating of reviews, in contrast, only depends on the
peer reviews and the authors responses.
%\todo{YG: address the question in review about why we do 
%not predict final acc/rej decisions.}

We choose a classic feature-rich classification 
model for this task, for 
%three 
two reasons: 
a) model capacity is lower compared to, e.g., a deep 
neural network, which is beneficial in our small 
data scenario,
%that we are facing,
and
b) the results are easier to interpret.
%
%We exclude features that are not accessible
%to authors,\todo{SE: to the reviewers? YG: no, to %authors. We are aiming to help authors to predict the %final score, right? SE: I see, but on the other hand %we also model, in general, what causes score updates, %so I'm not sure if we should motivative it like this. %But if so, we could write "our target group, authors, %"} e.g., submission time of reviews. 

\subsection{Features}
%Suppose we try to predict the score update of review $i$ for
%submission $p$. Below we discuss the score-based
%and text-based features we used.

\noindent \textbf{Score features (\texttt{Score})}.
We use all peer review scores for a given submission 
%$p$ 
to build %the 
an array of 
score-based features.
%\todo{SE: @Alex, stop using genetive, it's not common in modern English. It identifies us as Chinese :P}
These include review $i$'s before-rebuttal \os{} 
(\texttt{self\_score}), % and confidence score (\texttt{self\_conf}),
statistics of the other peer reviews' \os{} (denoted by \texttt{oth\_X}, where \texttt{X} can be 
\texttt{max}/\texttt{min}/\texttt{mean}/\texttt{median}/\texttt{std}),
statistics of all peer reviews' \os{}
(\texttt{all\_X}),
and elementary arithmetic operations on the above features
(e.g., \texttt{oth\_mean-self} denotes the mean \os{} of the peer reviews
%overall score 
minus review $i$'s before-rebuttal \os{}).
%Confidence scores 
\texttt{CONF}
%and other aspect scores are also considered similarly.
are considered in a similar manner. We do not consider aspect scores such as \texttt{ORG} because 
they yielded no improvements in our preliminary experiments. 
The full list of features can be found in the supplementary material.
%
%Next, we discuss our text-based features. 
We also include features based on the author response texts,
as detailed below. 

\noindent \textbf{Length of response (\texttt{log\_leng})}.
%Good author responses are usually longer than 
%normal and bad ones 
We have found that high-quality author responses
are usually longer than the low-quality ones
(see \S\ref{subsec:rebuttal_analyses}).
We use the logarithm of the number of tokens in author responses
as a feature.

%\YG{
%\noindent \textbf{Review submission time (\texttt{submit\_time})}.
%On average, the INC reviews are submitted significantly 
%earlier than the DEC reviews (see \S\ref{subsec:reviews_analyses}).
%We use the time gap (in hours) between the review submission time
%and the submission deadline as a feature.
%\todo{yg: in practice, authors cannot access this info. Not sure
%whether we need to add this feature.}
%}

\noindent \textbf{Review-Response Similarity (\texttt{sim})}.
Lack of similarity between a review and its response
may indicate that the response is ``off-topic''.
To measure similarity, we have trained 300-dimensional skip-gram word embeddings on 
5611 papers extracted from the cs.CL (computational and language) 
and cs.LG (learning) categories of ArXiv which were published 
between January 1, 2015 and December 31, 2017. 
%\todo{YG: from when? starting time? SE: good question. I'm not sure I can find this out easily. YG: OK, I just use a fake date as placeholder in
%the draft.}
We represent reviews and responses
by averaging the embeddings of their words, and measure 
semantic similarity by cosine similarity.\footnote{We also used ROUGE \cite{lin2004rouge} to measure the similarity 
but find the ROUGE scores to be highly correlated with the cosine similarities (Pearson correlation
$>$ 0.9), so we include only the cosine similarities in our models.}
 %\todo{SE: ROUGE footnote may be dropped}
%We find that the similarity measured with ROUGE-1, ROUGE-2 and
%average embedding cosine are very similar (pearson correlation
%$>0.95$ between any pair of measure), hence we only
%use the embedding cosine similarity in our regression.}
We find it important to use word embeddings 
trained on \emph{CL/LG} domain data: 
%as more popular embedding techniques such as InferSent \cite{Conneau:2017}, trained on general domains, yielded representations that could not sufficiently discriminate between scientific texts: 
%either due to a high number of out-of-vocabulary words or due to unspecific representations of scientific terms resulting from domain shift. %\footnote{E.g., 
for example, nearest neighbors of ``neural'' in a model trained on Wikipedia are ``axonal'', ``salience'', 
while on Arxiv its nearest neighbors are
%while nearest neighbors for our model trained on Arxiv are 
``feedforward'' and ``deep''.
We find that \ir{}s are more similar
to their reviews than \dr{}s and \kr{}s:
the average cosine similarity between the reviews and 
\ir{}s, \dr{}s and \kr{}s are .38, .30 and .29, respectively. 
%see the appendix, Figure \ref{fig:}, for plots of distributions.
    
\noindent \textbf{Specificity (\texttt{spec})}.
%Responses that are more specific can more effectively 
%address the questions and criticisms in reviews, 
%and thus are more likely to be good.
%As we found 
In our human annotation experiments, %, that
unspecific responses were typically judged as weak 
because they did not address specific questions or 
weaknesses given by reviews. To measure
the specificity of author responses, we use a feature-rich sentence-level specificity model
by \newcite{li2015fast} trained
on multiple news corpora. The produced scores
are in the $[0, 1]$ range, with higher values meaning higher specificity.
%uses sentence surface features,
%dictionary features and lexical features. It is trained
%on multiple news corpora. 
%To validate its performance on the author responses, 
%we randomly selected 15 pairs of sentences from 
%different author responses and asked the same three %experienced 
%reviewers as before to indicate the more 
%specific sentence in each pair.
%We find 87\% of the annotations agree, and the aggregated
%annotation (by majority voting) agrees with the score-induced
%preferences in 93\% cases,
%suggesting that the specificity scores are good. 
%Example sentences and scores
%are presented in the the supplementary material.
\ir{}s are slightly more specific than the other responses:
the mean specificity scores for
%for sentences in 
\ir{}s, \dr{}s and \kr{}s are 
.29, .24 and .28, respectively.
%further suggesting that good responses are indeed more specific.
For each author response, we
compute the \texttt{spec} scores for all their sentences
and use statistics 
(\texttt{max}/\texttt{min}/\texttt{mean}/\texttt{median}/\texttt{std}) of the \texttt{spec} scores as features.
The same strategy is used to build the
politeness and convincingness features introduced below.
    
\noindent \textbf{Politeness (\texttt{plt})}.
%Politeness is a key component of 
%human communication, and the interaction between reviewers 
%and authors is no exception, although a polite rebuttal does not 
%always lead to a score improvement. 
We employ the sentence-level politeness framework
suggested by \newcite{politeness} to quantify the politeness
of the author responses. We have trained a simple bag-of-words based multi-layer perceptron (MLP) model 
using their Wikipedia and StackExchange data and applied it 
to the author responses, generating a politeness score in $[-1, 1]$ 
for each sentence in author responses, where higher scores
mean higher politeness. 
%Despite the model simplicity and the domain shift, 
%Our manual analysis reveals that the model assigns reasonable politeness scores.
%To validate the correctness of the politeness scores,
%we again performed the pairwise comparison experiments
%on 15 randomly selected sentences with 3 experienced reviewers.
%The agreement between the annotators is 0.87,
%and the agreement between the score-induced preferences
%and the aggregated annotators' preferences is 0.87,
%suggesting that the politeness scores can differentiate 
%the 
%polite and non-polite sentences in author responses.
%e.g. \textit{"Thank you for a great suggestion"} 
%(0.75), \textit{"This will be clarified in camera-ready"} (0.12), 
%\textit{"What's not clear about our evaluation setup?} (-0.44). 
%Interestingly, we observe that the rebuttals leading to a score increase 
%are on average \textit{less} polite than score-decrease and no-change 
%rebuttals (0.18, 0.19 and 0.20 respectively).
While the mean politeness scores in \ir{}s, \dr{}s 
and \kr{}s  have no marked differences 
(all around 0.19), the 
score for the most polite sentence in 
\ir{}s (.91) is higher than 
that of \dr{}s (.68) and \kr{}s (.90).

\noindent \textbf{Convincingness (\texttt{cvc})}.
To approximate rebuttal convincingness we use the sentence-level convincingness model 
developed by \newcite{simpson2018tacl}, trained on $\sim$1.2k argument
pairs from web debate forums.
%We again performed the pairwise comparison experiments on
%15 randomly selected sentences.
%The agreement between 3 annotators is 0.64, 
%and the agreement between the aggregated annotators'
%preferences and the score preferences is 0.67.
%Note the agreement between annotators is lower than
%in Politeness and Specificity evaluations,
%suggesting that measuring the convincingness is even
%a hard task for experienced reviewers.
We normalize all convincingness scores to $[0,1]$,
where larger scores mean higher convincingness.
Mean convincingness scores for \ir{}s, \dr{}s and \kr{}s 
are .60, .49 and .58, respectively.
%, suggesting that
%good reviews are indeed more convincing.
%\todo{SE: the feature description section is (way) too long}

\noindent \textbf{Score validation}.
Since the \texttt{spec}, \texttt{plt} and \texttt{cvc} models
are not trained on review-rebuttal data, %author response data, 
%and we are thus facing a domain shift, we perform
we need to perform
human evaluations to validate the produced scores.
We rank the sentences in author responses
in terms of their \texttt{spec},
\texttt{plt} and \texttt{cvc} scores and analyze the top and bottom 10 sentences in each ranking
(see the supplementary material).
We find that the scores successfully distinguish the most
and least specific/polite/convincing sentences.
To further validate the scores,
for each type of score, we have randomly sampled 15 pairs 
%\todo{SE: reviewers might say this is very little; YG: maybe do more}
of sentences from author responses and presented the 
pairs to 3 experienced annotators, asking them
to indicate the more specific/polite/convincing sentence
in each pair. The agreement is presented in Table \ref{tab:agreement}.
The agreement between the users' aggregated preferences
and score-induced preferences is quite high for 
all three types, confirming the validity of the scores.
Note that the agreement for \texttt{cvc} is lower
than the other two; 
%the reason may be 
%higher discrepancy between the domain the classifier has been trained on and the domain it is used for. \todo{SE: should we remove that? Because user disagreement is not related to domain shift ...} Alternatively, 
the reason might be that it is difficult even for humans to judge convincingness of arguments, particularly when evaluated on the sentence level without surrounding context nor the corresponding review.
%\todo{SE: this may be used as an attack against us.} 
%that, when only presented
%with a sentence without its context and corresponding review,
%it is difficult to precisely measure 
%even experienced humans have difficulty to indicate 
%the convincingness
%even for experienced reviewers.
%Example responses and their \texttt{spec}/\texttt{plt}/\texttt{cvc}scores, and 
The distribution of the \texttt{spec},
\texttt{plt} and \texttt{cvc}
scores for \ir{}s, \dr{}s and \kr{}s is %can be found 
in the supplementary material.

%We compute 
%\texttt{spec}, \texttt{plt} and \texttt{cvc}
%on sentence level for each
%author response, and use
%their statistics (e.g., max/min/mean)
%as features. See the supplementary material
%for the full list of features.

\begin{table}[]
    \small
    \centering
    \begin{tabular}{l|l l l}
        \toprule
         & \texttt{spec} & \texttt{plt} & \texttt{cvc} \\
         \midrule
         Inter-User & .87 & .87 & .64   \\
         User-Score & .93 & .87 & .67 \\
         \bottomrule
    \end{tabular}
    \caption{Percentage of agreement for \texttt{spec}, \texttt{plt} 
    and \texttt{cvc} scores. ``User-Score'' means the agreement
    between the aggregated (by majority voting) users' preferences 
    and score-induced preferences.}
    \label{tab:agreement}
\end{table}

\subsection{Results and Analyses}
\label{subsec:results_analyses}
We perform %our 
experiments on a subset
of the corpus which only includes the submissions
that have author responses and three or more reviews opted in.
We term this subset of the corpus 
\emph{Submissions with Complete Reviews
(\full)}.
Training models on submissions with fewer reviews
would bias certain features (e.g. \texttt{all\_mean})
and thus bias the trained models.
Also, we separate out the submissions from the Full set
whose before-rebuttal average \os{} are between 
3 and 4.5 (note that \os{} are in $[1,6$]), 
so as to train and test a model
specifically on borderline submissions for which score 
changes may be decisive for an accept or reject decision.
We term this subset  \emph{Borderline Submissions (\border{})}.
%\todo{SE: that's a misleading name}
%We perform experiments on SCR because
%in the conference 
%most papers receive 3 reviews and author responses; 
%ignoring some reviews for a paper would mean 
%to ignore important influence variables.
\full{} includes 791 submissions (80 INC, 60 DEC, 652 KEEP)
and \border{} includes 590 (69 INC, 48 DEC and 474 KEEP).
All results and weights presented in this section 
are averaged over 5000 repeats of 
10-fold cross validation;
data entries are randomly shuffled for each repeat.

\paragraph{Feature Selection.} We filter out features whose information gain
is ranked in the bottom 50\% of all features on the training set.
%\todo{SE: on the train set?} %and the highly correlated
For highly correlated features in the upper 50\%  (i.e.\ Pearson correlation $\geq 0.5$),
we filter out all but the one with the highest information gain. 
%\todo{SE: if two features are highly correlated, which one is kept?}
Remaining features are used to train a
%and use the remaining features to train a 
multinomial logistic regression model (i.e., 
MLP with no hidden layer and softmax activation 
function in the output layer). 
To balance the number of instances for the three classes,
on the training set, in each fold
of cross-validation we randomly down-sample cases with class KEEP to 
ensure that the number of KEEP is the same as the sum
of INC and DEC.
We also tried random forest, decision tree,
support vector machines and Gaussian processes as  
classifiers, but their performances were similar or %even 
worse than that of logistic regression.

\paragraph{Results.}
Classification results are presented in Table \ref{table:test_results}. In addition, we compare to two baselines: the \emph{majority baseline} always picks the majority decision (in our case, KEEP); the \emph{random baseline} selects an action at random.
Full results, including precision, recall and F1-scores
for each label, can be found in the supplementary material.

We find that score-based features are most effective 
among all features. 
%But the 
However, text-based features are also useful,
supported by the observations that:
\textbf{(i)} models using only text features all 
significantly (p-value $<0.01$, double-tailed t-test)
%\footnote{%Unless stated otherwise, 
%in this paper, 
%We use double-tail t-test to 
%compute p-values, and %let significance level 
%choose the significance level as $0.01$.} 
outperform the majority and random baseline; and 
%\textbf{(ii)} using score-based features with a single text-based
%feature set results in significant improvement,
%except for \texttt{Score+plt}; and
%which even performs worse than
%using Score features alone; and
%\todo{YG: reason is unclear. Score and Plt have quite low
%correlation ($<$0.2).}
\textbf{(ii)} using all features gives the best performance, 
significantly (p-value $<0.01$) better than using any feature set alone.

%\texttt{log\_leng} is the most effective text-based 
%feature in both BDS and SCR when being used alone.
%Taking the score Length is one of the best
%Aside from the \texttt{Score} features, 

Among the non-\texttt{Score} features, 
%Taken alone, 
\texttt{log\_leng} performs best. But we
find it has high correlation with
multiple \texttt{Score} features, and hence
when all features are used, it %\texttt{log\_leng}
is filtered out. % (see the analysis in the paragraph below).
%is the best non-\texttt{Score} feature, and it is relatively better in \border{} than in \full{}. 
%Note that 
The features \texttt{spec} and \texttt{sim} 
perform much better in \border{} 
than in \full{},
%while \texttt{plt} performs 
%better in \full{} than in \border{}. 
which suggests that, 
%in general, reviewers
%consider politeness and the length of the author response
%when they decide whether to change their scores,
%important when deciding whether to change their scores,
%but 
for borderline papers, more weight is placed 
on whether the response explicitly addresses
the points raised in reviews (similarity)
and the specificity of the response.%,
%while in generic cases, 
%\todo{SE: politeness matters ... does that not contradict with below?}

%\todo{SE: we might drop this} 
%We have also tried other feature selection techniques
%(principle component analysis with exhaustive dimension search) 
%and other classifiers 
%(one-vs-rest logistic regression, random forest, 
%decision tree, SVM with polynomial, linear
%and RBF kernel, and Gaussian process),
%but %no significant improvement is observed.
%these showed no better results than multinomial logistic regression.
%\todo{YG: ranking of features by different %algorithms
%is missing now. Do we need it? SE: I would like to %see it, if possible.}

\begin{table}[t]
    \small
    \centering
    \begin{tabular}{r|ll}
        \toprule 
         Feature Set & \border{} & \full{} \\ \midrule
         \texttt{spec} & .324 & .309 \\
         \texttt{plt}& .306 & .310 \\
         \texttt{cvc} & .303  & .304 \\
         \texttt{log\_leng}& .340 & .341 \\
         \texttt{sim} & .323  & .302 \\
         \texttt{Score} & \textbf{.495} &  \textbf{.526} \\ 
         \midrule
         %\texttt{Score+spec} & 0.510 & 0.529 \\
         %\texttt{Score+plt} & 0.491 & 0.521\\
         %\texttt{Score+cvc} & 0.497 & 0.532 \\
         %\texttt{Score+log\_leng} & 0.505 & 0.530 \\
         %\texttt{Score+sim} & 0.516 & 0.528 \\
         %\midrule
         All but \texttt{Score} & .343  & .336 \\
         All & \textbf{.522} & \textbf{.540} \\  
         \midrule
         Majority Baseline & .297 & .301 \\ 
         Random Baseline & .258 & .251 \\
         \bottomrule
    \end{tabular}
    \caption{Macro F-1 scores. 
    %All results are averaged
    %over 5000 repeats of 10-fold cross validation.
    }
    \label{table:test_results}
\end{table}
%\todo{SE: does not converge to the confident person; also, conf. does not play any role in the classification. This conflicts with existing studies on humanities. What we observe is only the peer pressure. }

\paragraph{Analysis.}
%We also study 
To interpret our results, 
we study the weights of the features in our logistic regression model  
%the multinomial logistic regression.
%Weights for the model trained on SCR and BDS are presented 
%
shown in Tables \ref{tab:weights_scr} and \ref{tab:weights_bds}.
%,
%respectively. We interpret the weights as follows. 
We observe the following trends: 
\begin{itemize}
    \item \textbf{``Peer pressure'' %has the strongest influence on
    is the most important factor of score change}:
    in both \full{} and \border{}, features reflecting the gap between 
    own and others' review scores (\texttt{oth\_mean-self} and \texttt{self-oth\_min}) 
    have by far the largest weights compared to other feature groups. For example, in \full{}, the \texttt{Score} features have (absolute) weights of 0.4 or higher for the class INC, while all other features are substantially below 0.2. The weights make intuitive sense: e.g., when the mean of the other reviewers' scores is above a reviewer's initial score, she has a strong tendency to increase her own score and not to decrease her own score. Similarly, when a review contains a very convincing sentence, this substantially decreases the probability of a score decrease. 
    \item \textbf{To improve the score for a borderline paper,
    %you need 
    a more convincing, specific and explicit response may be helpful}: 
    in \full{}, no weight of a text-based feature is above 0.2 for INC;
    however, in \border{}, the weights for \texttt{cvc\_min}, \texttt{spec\_median} and \texttt{sim}
    are all above 0.2. 
    %More generally, weights for text-based features are typically considerably larger in \border{} compared to \full{}. 
    This asymmetry of the text-based features across \full{} and \border{} also suggests that 
    %as a surprisingly positive outcome of the rebuttal phase, 
    reviewers do appear to pay more attention to the author responses in situations where they may matter (e.g., make the difference between accept or reject decisions). 
    \item \textbf{An impolite author response may harm the final score}:
    in both \full{} and \border{}, the weight of \texttt{plt\_max} is negative for DEC.
    In addition, %note that 
    in \full{} a more polite response helps increase the final score
    (positive weight for INC, close to 0 weight for KEEP).
    %while 
    In \border{}, in contrast, a more polite response may not increase the score but only
    keep it unchanged (positive weight for KEEP, close to 0 weight for INC). If we take \border{} papers as those for which the author responses really matter, this means that politeness has an asymmetrical effect: it may push a paper below the acceptance threshold, but not above it. Indeed, \texttt{plt\_max} is the second best text-feature for predicting decrease for \border{} papers. 
    %suggesting
    %that response with high \texttt{plt\_max} is more likely 
    %to result in keeping the score, but those with 
    %low \texttt{plt\_max} can result in score drop.
\end{itemize}

\begin{table}[]
    \small
    \centering
    \begin{tabular}{r | c c c}
        \toprule
         Feature & INC & DEC & KEEP \\
         \midrule
         \texttt{oth\_mean-self} & 1.044 & -1.265 & .221  \\
         \texttt{self-oth\_min} & -.378 & .188 & .190 \\
         \texttt{cvc\_max} & .078 & -.271 & .193 \\
         \texttt{spec\_median} & .159 & -.224 & -.065 \\
         \texttt{plt\_max} & .170 & -.174 & .004 \\
         \texttt{sim} & .019 & .099 & -.119 \\
         \texttt{spec\_max} & .022 & .029 & -.051 \\
         \bottomrule
    \end{tabular}
    \caption{Feature weights in multinomial logistic regression trained on \full{}. 
    %All weights are averaged over 5000 repeats of 10-fold cross validation.
    }
    \label{tab:weights_scr}
\end{table}

\begin{table}[]
    \small
    \centering
    \begin{tabular}{r | c c c}
        \toprule
         Feature & INC & DEC & KEEP \\
         \midrule
         \texttt{oth\_mean-self} & .855 & -1.026 & .171  \\
         \texttt{self-oth\_min} & -.372 & .191 & .181 \\
         \texttt{cvc\_min} & .224 & -.258 & -.034 \\
         \texttt{spec\_median} & .293 & -.122 & -.171 \\
         \texttt{sim} & .214 & -.161  & -.053 \\
         \texttt{cvc\_max} & .117 & -.085 & -.033 \\
         \texttt{plt\_max} & .016 & -.192 & .176 \\
         \bottomrule
    \end{tabular}
    \caption{Feature weights in multinomial logistic regression trained on \border{}. 
    %All weights are averaged over 5000 repeats of 10-fold cross validataion.
    }
    \label{tab:weights_bds}
\end{table}

\label{sec:scorechange}
\section{Discussion}
%Why are the reviewers score-update decisions
%so strongly influenced by the ``peer pressure''?
%Is the influence of ``peer pressure'' also observed
%in other group-opinion-formation processes besides %scientific
%peer review? In this section, 
%We now relate our observations
%to studies in %economics and human behavioural, 
%so as to better understand the root of the peer 
%pressure effect.\todo{SE: ...}
%\todo{SE: I removed the introductory words. Even though Nafise may not have seen the relationships, I do think that they are quite obvious}
The opinion update process we have described in \S\ref{sec:scorechange} is closely related to the work 
%on %so-called 
on \emph{opinion dynamics} \cite{DeGroot:1974,Acemoglu:2011}, 
which studies how human subjects change their opinions %based on their peers' opinions.
as a reaction to those of peers. 

The ``peer pressure'' effect (opinions being updated to mean opinions) is widely observed
in opinion formation of human subjects 
in controlled experiments. 
%\citet{Lorenz:2011} find that social influence has a negative effect on crowd-wisdom in simple estimation tasks (``What's the population density of Switzerland?''): human subjects tend to lean towards a consensus once they are exposed to the opinions of others, %which does not necessarily lead to a better estimate of the true value. 
%\YG{
\citet{Lorenz:2011} find that in 
simple estimation tasks (``What's the population density of Switzerland?''),
human subjects tend to lean towards a
consensus once they are exposed to the opinions of others.
%\SE{without this leading to a better estimate of the true value}.
%but this consensus often does not help the group to make
%a better estimate of the true value.
%}
%Putting their findings in our peer
%review context, we believe that the peer pressure %influence is
%inevitable if the reviewers can see other peer %reviews.
%If we limit the visibility of peer reviews, e.g.
%not allowing reviewers to see peer reviewers' %scores,
%the peer pressure effect might be reduced.
Similarly, \citet{Moussaid:2013} find two dominant effects for simple factual questions: human subjects tend towards the mean opinion and towards the opinions of highly confident individuals. Our experiments 
%concur with the last two controlled experiments in that 
also show that the 
mean opinion plays a very prominent role in peer reviews,
%\YG{
but they show %Interestingly, we find 
no evidence supporting the confidence effect: 
features based on the confidence scores do not
play a significant role in deciding the final
scores (see \S\ref{sec:scorechange}).
%we do not observe any strong correlations between
%the influence of a reviewer \todo{SE: how's the influence of a reviewer defined?} and her confidence score
%or her seniority level \todo{SE: wow, where does seniority come from? appendix?}.
%
We believe this is due to two main differences
between peer reviewing and the controlled experiments
in the above works:
(i) there does not exist a ground-truth score
for a submission, while such true answers 
about factual questions do exist in 
the controlled experiments;  and (ii)
%unlike the participants of controlled experiments 
%with \emph{monetary stakes},
participants of the controlled experiments lose money
if they give incorrect answers, but
a reviewer loses nothing %even if 
when 
she does not 
adjust %her reviews 
to a (self-assessed) expert.
%or a more senior reviewer. 
%}

%and a \emph{pre-defined ground truth}, 
%, and additionaly,.
%from not adjusting 
%to a (self-assessed) expert, as wrongly estimating the quality of the submission is free of charge, and a true score for a submitted paper does not exist in the same sense as for factual questions. 
%might not even exist after all. 

%Research on opinion dynamics identified 
Three types of \emph{biases} (a.o.) %are widely observed in
have been studied in explanatory models of opinion dynamics in recent years.
The first is %an %result of 
opposition between
members of different groups (e.g., due to \emph{group-identity}) 
leading to distancing from certain subjects' opinions \cite{Altafini:2013,Eger:2016}.
%Even though it is not impossible that this bias exists in our data in individual cases, 
The second is \emph{homophily}:
%observed in situations where 
individuals ignore opinions
too different from their own \cite{Deffuant:2000,Hegselmann:2002}. 
%We found no direct evidence for these two types of bias 
%on an aggregate level, as the mean opinion is a 
%strong attractor for individual opinions. 
The third is \emph{conformity}
\cite{Buechel:2015}, i.e., %the disutility
%negative feelings  
%experienced by individuals when not conforming 
the desire to conform 
to a group norm/opinion. 
%happened when xxx\todo{@SE, a bit more explaination here?}. 
Conformity bias can be strong and 
persist even in the presence of %very strong
overwhelming 
evidence that a group opinion is wrong \cite{Asch:1951}.
Our observation that reviewers tend to converge
to the mean of all reviews
(\S\ref{subsec:results_analyses}) suggests that
%the conformity bias is likely to be the primary
%source of bias in the rebuttal process.
conformity bias %may 
also plays a prominent role in peer reviewing.
We found no evidence (on an aggregate level) for the other two biases.

To summarize, conformity bias is the %strongest 
main 
bias
we %observed 
identified 
in the peer reviewing process. 
However, conformity bias 
has a negative effect on crowd-wisdom in
estimation tasks \cite{Lorenz:2011},
%because it 
which 
strengthens %human subjects' 
confidence of human subjects in the
correctness of their converged answer, 
%but 
while 
the actual correctness of %their converged answer 
their consensus 
is often even worse than the mean of multiple
independent answers.
%
%We believe this may partly explain why peer-reviewing %fails 
%to identify major flaws in papers and %to select
%%the 
%papers with high future citations %in future
%(see \S\ref{sec:related}).
%
A simple method to reduce %the 
conformity bias
is to blind reviewers from each other, only
allowing reviewers to update their reviews based on
the author responses; the area chair (who can see
all reviews for a paper) is then responsible for considering
all (possibly conflicting) reviews and making 
the accept/reject recommendation.
%
%Apparently, improving the effectiveness of %peer-reviewing
%requires community-wide discussion and consensus,
%even collaborations with other communities such as
%researchers from opinion dynamics.
We believe that peer reviewing is to a large degree an opinion dynamics process, a neglected insight hitherto, and that lessons from this field should therefore be beneficial for peer reviewing for NLP conferences and beyond. 

Finally, concerning the helpfulness of individual 
review based feature groups, 
%\YG{
we believe it reflects a weakness of the current
rebuttal stage that%}
%we perceive it as a negative outcome that 
politeness does matter, because this is \emph{merely} a social %phenomenon 
aspect 
%that 
%has nothing to do with 
%is 
unrelated to the quality of the assessed papers. 
However, 
%contrary to common belief,\footnote{In which blog was this? } 
%\todo{YG: I don't think common belief is that filling up
%thank you can be very helpful. I think removing it is %fine.}
we also showed that filling up author responses with ``thank you''s is unlikely to increase a reviewer's score for a borderline paper---so at least, 
%there appears to be no selection bias 
authors do not seem to be able to sneak their papers  in %by being friendly. 
via social effects. 

\label{sec:discussion}
\section{Conclusion}
%In this paper, 
We presented a review corpus consisting of
over 4k reviews and 1.2k author responses from 
\acl{}.
%a major NLP conference.
To the best of our knowledge, it is the first corpus
that includes both before- and after-rebuttal
reviews for both accepted and rejected papers
in a major NLP conference.
We %performed thorough 
qualitatively and quantitatively analyzed  
%on 
the corpus, including a manual classification of 
%prominent 
paper weaknesses outlined by reviewers %of some submissions
%as well as the convincingness of some author responses.
and a %convincingness 
quality rating study of the corresponding author responses. 
%so as to help the community better understand the reviewing process.

In addition, 
%to quantitatively measure the 
%influence of author responses in the
%score-update decisions,
%we proposed a novel \emph{after-rebuttal score change prediction}
%task on the corpus and developed %a feature-rich 
we proposed a classification model to predict
whether a reviewer will increase/decrease/keep
her overall score after rebuttal.
By analyzing the feature weights %of features in our classification
in our 
model, we quantitatively measured the importance of different %factors
decision variables for %the %after-rebuttal 
score updates. % decisions.
%Interestingly, 
We %have found 
found that
the gap between %a review's and its peer reviews' before-rebuttal
%overall scores 
a reviewer's initial score and her peers' scores 
is the main explanatory variable. 
%factor %that affects the 
%explaining the score updates. 
%while the effects of some other
Rebuttal-related %response-related 
factors like convincingness, %of the response,
specificity and politeness of %author 
responses are %less 
considerably less important but still have a
statistically significant effect, especially for borderline papers.\footnote{We believe that they might become more important when further argumentation/text-based features are integrated.}
Our findings shed light on the predominant role 
of the \emph{conformity bias} in peer reviewing
(see \S\ref{sec:discussion}),
%an oft-heard but never quantitatively measured
%bias in peer-reviewing, 
and we discuss alternative peer review
models %combating 
addressing 
this bias. We hope our
%findings and discussions 
analyses 
will help the community better
understand the strengths and weaknesses of the current 
peer review workflow, %and 
%hence encourage more discussions
%and studies in this subject.
%hope they will 
spurring further discussions. %studies. 
%\todo{YG: I remove the discussion of the usefulness %of
%rebuttal here, to highlight our findings on the %biases.}

%However, inferring from our results that the rebuttal phase is 
%of negligible value overall 
%would be premature, since our features were 
%hand-selected (which runs the risk of missing 
%important information signals) and obtained from 
%%classifiers trained on different domains 
%such as web data (see \S\ref{sec:score_classification}); 
%in addition, our dataset for training the feature-based 
%models has still been comparatively small.
%
%Nevertheless, we hope that our findings 
%\YG{and our discussions on the biases in 
%decision-making process of reviewers (see \S\ref{sec:discussion})
%can help the community better understand and even
%improve the peer-reviewing workflow in NLP conferences.}
%shed further light on the 
%decision-making process in major NLP conferences 
%regarding score updates in the rebuttal stage, 
%thus helping the community to re-evaluate its value. 

Finally,  provided that the rebuttal phase remains 
a key feature in many peer reviewed conferences, 
%of future NLP conferences
we think that our novel after-rebuttal score 
change prediction task can be practically beneficial for authors 
to restructure their author responses and thereby make them more effective. \label{sec:conclusion}
\section*{Acknowledgement}
The authors thank the anonymous reviewers and Dan Jurafsky  for %their 
constructive comments and %Dan Jurafsky for his 
helpful remarks.
This work has been supported by the ArguAna Project GU 798/20-1 (DFG),
 the German Research Foundation through the 
 German-Israeli Project Cooperation 
 (DIP, grant DA 1600/1-1 and grant GU 798/17-1), 
and the DFG-funded research training group 
``Adaptive Preparation of Information from Heterogeneous Sources'' 
(AIPHES, GRK 1994/1).
Ilia Kuznetsov has been supported by the FAZIT Foundation.
\label{sec:ack}
\bibliography{naaclhlt2019.bib}
\bibliographystyle{acl_natbib}

\newpage
$ $  
\newpage
\section*{Supplementary Material}

\section*{Consent Message}
Before a reviewer or an author enters her reviews or author
responses, the following message appears to ask for her
consent for data sharing:

\begin{displayquote}
ATTENTION: this time, we plan to do some analytics on anonymized reviews and rebuttal statements, upon the agreement of the reviewers and authors, with the purpose of improving the quality of reviews. The data will be compiled into a unique corpus, which we potentially envisage as a great resource for NLP, e.g. for sentiment analysis and argumentation mining, and made available to the community properly anonymized at earliest in 2 years. We hope to provide data on "how to review" to younger researchers, and improve transparency of the reviewing process in ACL in general.

By default, you agree that your anonymised rebuttal statement can be freely used for research purposes and published under an appropriate open-source license within at earliest 2 years from the acceptance deadline.

Place an 'x' mark in the NO box if you would like to opt out of the data collection. \\
     %=X=0   
     \texttt{[x]}: YES\\
     %=X=1   
     \texttt{[ ]}: NO
\end{displayquote}

\section*{Analyses on Submissions}
\label{subsec:submissions}

We rank n-grams in both accepted and rejected papers according to the \emph{log-likelihood ratio} (LLR) statistic, %\cite{Dunning:1993},
taking both accepted and rejected papers as one big corpus, respectively. The goal is to find n-grams that occur unusually frequently in one of the two groups, relative to the respective other. 

%and 
Table \ref{table:ngrams} shows a few hand-selected n-grams with highest LLR for \emph{accepted papers};
high-LLR n-grams for rejected papers are not presented
due to licensing.
%(we omit rejected papers because we have no permission to publish those). 
To filter out noise, we only include n-grams that occur in at least 7 different papers. 
We can observe some interesting patterns: 
%We find a few interesting tendencies of what the NLP community currently collectively cares about: 
accepted papers appear to cite recent work, which reflects potential novelty and appropriate comparison to state of the art; tend to use more mathematics (of a particular kind); have an appendix; do significance testing; release code upon publication; and have multiple figures including subfigures. 
%\todo{YG: @SE, as suggested by Nafise, maybe to give reference
%to some n-grams. SE: I don't understand?}
%\todo{SE: who wrote SOTA preprocessing for Dozat and Manning? The only work of Dozat and Manning I see has the title: "Deep Biaffine Attention for Neural Dependency Parsing."}
%In contrast, rejected papers tend to make use of older techniques (SVMs, LDA), and in particular do not use deep learning (in an expert way), but instead rely on hand-crafted features.  
%\todo{SE: maybe remove the possible interpretations}
\begin{table}[!htb]
    \centering
    \footnotesize
    \begin{tabular}{ll}
         \toprule
         Hot n-grams & Possible Interpretation 
         %&  Bad n-grams & Possible Interpretation 
         \\ \midrule
         ( 2017 ) & Cite recent work \\
         %& chinese word segmentation & narrow topic \\ 
         ( $z|x$ ) & Math 
         % &  , $\psi$ ( & features, non-neural
         \\
         artetxe et al & Authors working on a hot topic 
         % & discourse relation recognition & 
         \\
         dozat and manning & %Modern method? \\
         Authors of an influential method \\
         %& support vector machine & Obsolete trend \\
         in the supplementary & Paper has appendix \\
         %& implicit discourse relation & \\
         contextualized word & Trendy method \\
         %& neural network ( & Experts would use an abbreviation?\\
         representations & \\
         %& & \\
         upon publication . & Code/data will be released \\
         % Code/Data published & latent dirichlet allocation & LDA not recent technique?\\
         statistical significance of & Mathematically rigorous \\
         %& firat et al. & \\
         %\S2 ) . & Fancy abbreviation? \\
         %& ( blei et & LDA\\
         figure 3 ( & {\small Multiple figures with subfigures} \\
         %& mikolov et al. & Probably referring to 2013 paper\\
         \bottomrule
    \end{tabular}
    \caption{Selected 3-grams that distinguish accepted from rejected papers based on the LLR statistics.}
    \label{table:ngrams}
\end{table}

\section*{Statistics on Reviewer Information}
In this section, we present some statistics of 
all 1440 reviewers of ACL-18.

\begin{figure*}
    \centering
    \includegraphics[width=0.9\textwidth]{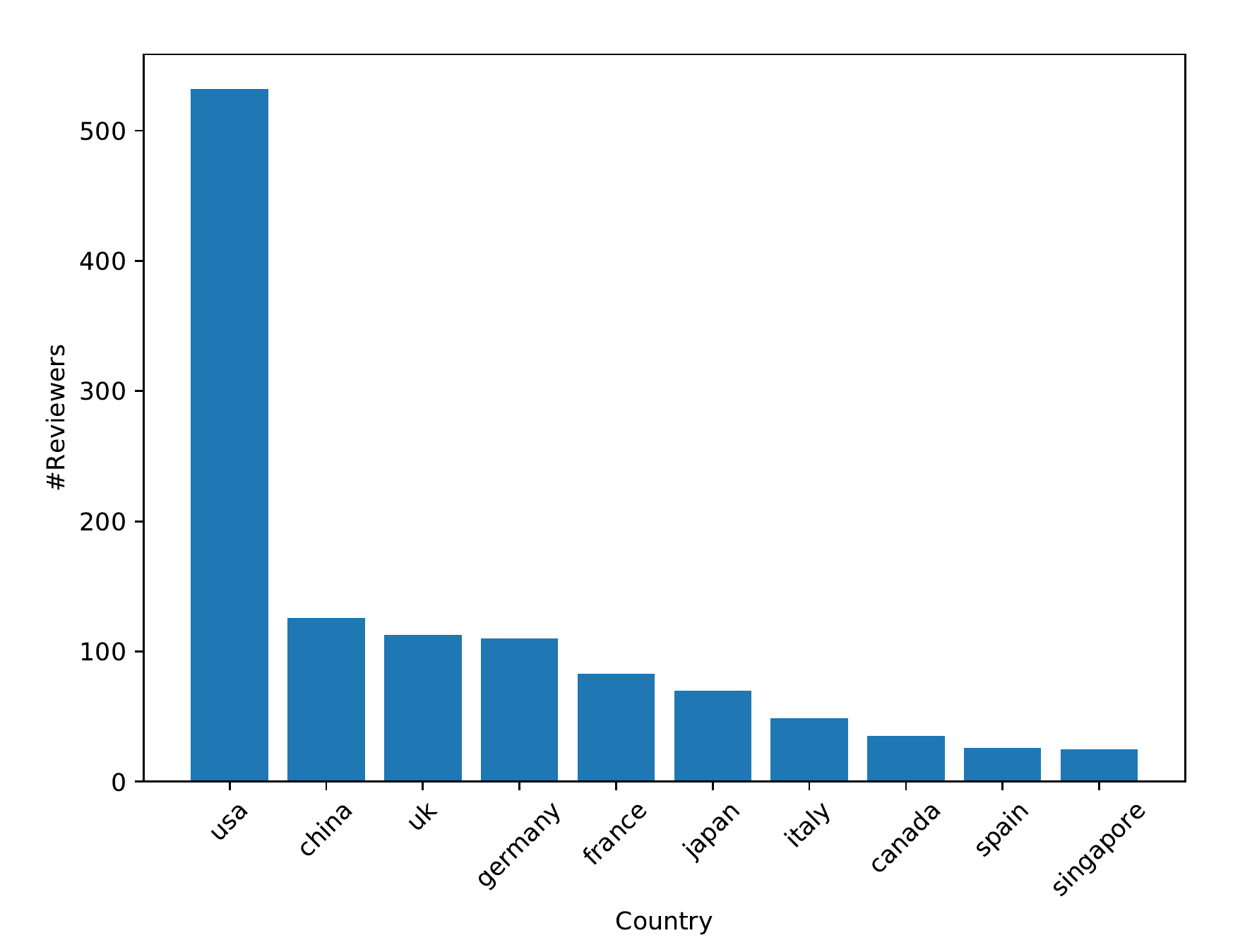}
    \caption{Distribution of countries where reviewers work.}
    \label{fig:country}
\end{figure*}

\paragraph{Country.} 
The reviewers work in 53 different countries.
The top 10 countries where the reviewers work
are presented in Fig.~\ref{fig:country}. 
The distribution of the reviewer working places 
is heavily long-tailed: the United States alone contributes 36.9\% of
all reviewers, followed by China (8.7\%), 
the United Kingdom (7.8\%) and Germany (7.6\%).
Seven countries have more than 50 reviewers,
and 19 countries have more than 10 reviewers.

\begin{figure*}
    \centering
    \includegraphics[width=0.9\textwidth]{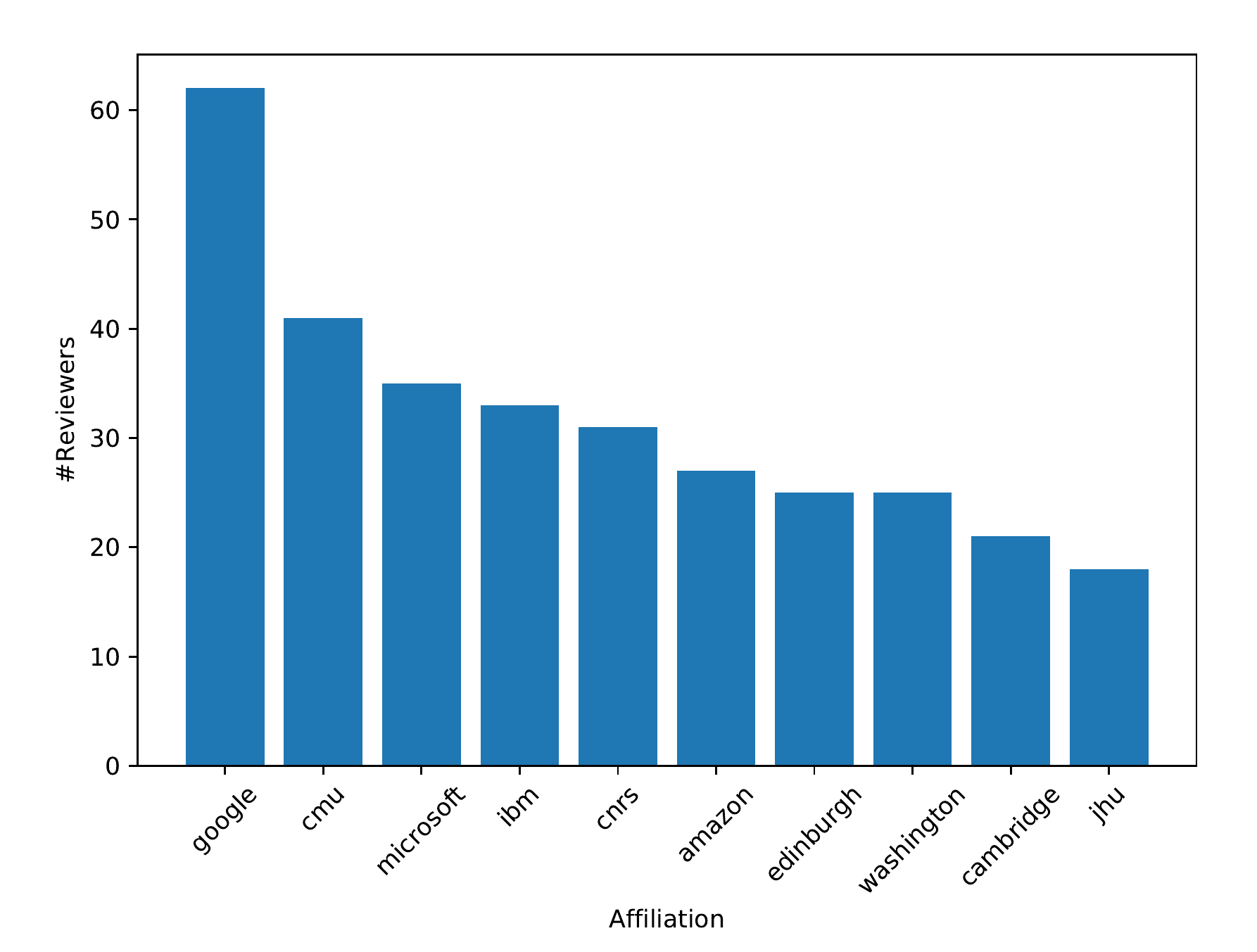}
    \caption{Distribution of organisations where reviewers work.}
    \label{fig:affiliation}
\end{figure*}

\paragraph{Affiliation.}
The reviewers are from around 700 organisations. 
But as reviewers use different names to 
refer to the same organisation
(e.g., both MIT and Massachusetts Institute of Technology are used),
%and are treated as different organisations),
the real number of organisations can be much lower. 
The top 10 organisations and their reviewers numbers are 
presented in Fig.~\ref{fig:affiliation}.
Nine organisations contribute more than 20 reviewers, and
19 organisations contribute more than 10 reviewers. %\todo{SE: where do all these numbers come from??} 

\paragraph{Seniority.}
Most reviewers (69.9\%) do not report their seniority levels.
Among those that have reported their seniority, 50.2\% are Professors,
27.6\% are PhD students, and 22.2\% are Post-Doc/Assistant-Professor.

\paragraph{Gender.}
We estimate the gender of the reviewers from their first names,
using the tool available at \url{https://github.com/kensk8er/chicksexer}. 
73.4\% reviewers are estimated to be male 
and the rest 26.6\% are estimated to be female.

\section*{Full Results}
The precision, recall and F1-scores
for each label in both \full{} and \border{}
are presented in Table \ref{table:full_results}
and \ref{table:borderlin_results}, respectively.

\begin{table*}
    \small
    \centering
    \begin{tabular}{r|ccccccccc}
        \toprule 
         Feature Set & INC-p & INC-r & INC-f1 &  
         DEC-p & DEC-r & DEC-f1 &
         KEEP-p & KEEP-r & KEEP-f1 
         \\ \midrule
         \texttt{spec} & .110 & .023 & .035 & 
         0 & 0 & 0 & 
         .820 & .976 & .892 \\
         \texttt{plt} & .043 & .063 & .047 & 
         .029 & .029 & .029 & 
         .824 & .912 & .864 \\
         \texttt{cvc} & .020 & .014 & .107 & 
         0& 0 & 0 & 
         .824 & .977 & .893\\
         \texttt{log\_leng} & .187 & .167 & .154 & 
         0& 0&0 & 
         .827& .930 & .874\\
         \texttt{sim} & .013 & .011 & .013 & 
         0 & 0 & 0 & 
         .810 & .990 & .897 \\
         \texttt{Score} & \textbf{.331} & .485 & \textbf{.386} & 
         \textbf{.380} & .527 & .409 & 
         .878 & .790 & .829 \\ 
         \midrule
         All but \texttt{Score} & .142 & .202 & .162 & 
         .025 & .033 & .029 & 
         .820 & .820 & .818 \\
         All & .299 & \textbf{.555} & .374 & 
         .364 & \textbf{.569} & \textbf{.438} &
         \textbf{.889} & .757 & .817\\  
         \midrule
         Majority Baseline & 0 & 0&0 &0 &0 &0 & 
         .823 & \textbf{1} & \textbf{.903} \\ 
         Random Baseline & .100 & .332 & .154 & 
         .076 & .334 & .123 & 
         .825 & .332 & .474 \\
         \bottomrule
    \end{tabular}
    \caption{Macro F-1 scores on \full. 
    All results are averaged
    over 5000 repeats of 10-fold cross validation.
    }
    \label{table:full_results}
\end{table*}

\begin{table*}
    \small
    \centering
    \begin{tabular}{r|ccccccccc}
        \toprule 
         Feature Set & INC-p & INC-r & INC-f1 &  
         DEC-p & DEC-r & DEC-f1 &
         KEEP-p & KEEP-r & KEEP-f1 
         \\ \midrule
         \texttt{spec} & .119 & .101 & .102 & 
         0& 0& 0& 
         .804 & .956 & .872  \\
         \texttt{plt} & .100 & .012 & .022 & 
         .020 & .014 & .017 & 
         .804 & .982 & .883 \\
         \texttt{cvc} & .033 & .020 & .025 & 
         0& 0& 0& 
         .803 & .988 & .885\\
         \texttt{log\_leng} & .180 & .229 & .184 & 
         0& 0& 0& 
         .811 & .879 & .840\\
         \texttt{sim} & .096 & .133 & .110 & 
         0 & 0& 0& 
         .805 & .927 & .861 \\
         \texttt{Score} & \textbf{.313} & .556 & \textbf{.394} & 
         .377 & .356 & .302 & 
         .851 & .743 & .792 \\ 
         \midrule
         All but \texttt{Score} & .205 & .331 & .231 & 
         .050 & .011 & .018 & 
         .801 & .768 & .780 \\
         All & .295 & \textbf{.570} & .376 &
         \textbf{.387} & \textbf{.548} & \textbf{.418} & 
         \textbf{.875} & .710 & .782\\  
         \midrule
         Majority Baseline &0 &0 &0 &0 &0 &0 & 
         .802 & \textbf{1} & \textbf{.890} \\ 
         Random Baseline & .117 & .333 & .173 & 
         .082 & .335 & .131 & 
         .802 & .333 & .470 \\
         \bottomrule
    \end{tabular}
    \caption{Macro F-1 scores on \border. 
    All results are averaged
    over 5000 repeats of 10-fold cross validation.
    }
    \label{table:borderlin_results}
\end{table*}

\section*{Features}
The full list of our hand-crafted features
is presented in Table \ref{tab:feature_list}.
%\todo{YG: change response type names in figures}

\begin{table*}[]
    \centering
    \begin{tabular}{l|p{12cm}}
        \toprule
         Feature set & Features \\
         \midrule
         \texttt{Score} & 
         \texttt{self\_before}, \texttt{self\_conf}, 
         \texttt{oth\_max}, \texttt{oth\_min}, 
         \texttt{oth\_mean}, \texttt{oth\_median}, \texttt{oth\_std}, 
         \texttt{oth\_conf\_max}, \texttt{oth\_conf\_min}, 
         \texttt{oth\_conf\_mean}, \texttt{oth\_conf\_median}, \texttt{oth\_conf\_std}, 
         \texttt{oth\_mean-self}, \texttt{oth\_median-self}, 
         \texttt{oth\_max-self}, \texttt{self-oth\_min}, 
         \texttt{oth\_conf\_std}, 
         \texttt{all\_max}, \texttt{all\_min}, 
         \texttt{all\_mean}, \texttt{all\_median}, \texttt{all\_std}, 
         \texttt{self\_before**2}, 
         \texttt{all\_mean-self}, \texttt{all\_max-self}, 
         \texttt{all\_median-self}, \texttt{self-all\_min}
         \\
         \texttt{spec} &
         \texttt{spec\_max}, \texttt{spec\_min}, 
         \texttt{spec\_mean}, \texttt{spec\_median}, \texttt{spec\_std}
         \\
         \texttt{cvc} &
         \texttt{cvc\_max}, \texttt{cvc\_min}, 
         \texttt{cvc\_mean}, \texttt{cvc\_median}, \texttt{cvc\_std}
         \\
         \texttt{plt} &
         \texttt{plt\_max}, \texttt{plt\_min}, 
         \texttt{plt\_mean}, \texttt{plt\_median}, \texttt{plt\_std}
         \\
         \texttt{log\_leng} &
         Logarithm of the token number of the author response
         %\\
         %\texttt{submit\_time} &
         %The time difference (in hours) between a review's 
         %submission time and the review submission deadline
         \\
         \texttt{sim} &
         Cosine similarity of the embeddings of a review and its
         corresponding author response \\
         \bottomrule
    \end{tabular}
    \caption{The full list of hand-crafted features.}
    \label{tab:feature_list}
\end{table*}

\section*{Specificity Scores}

\begin{figure}
    \centering
    \includegraphics[width=0.45\textwidth]{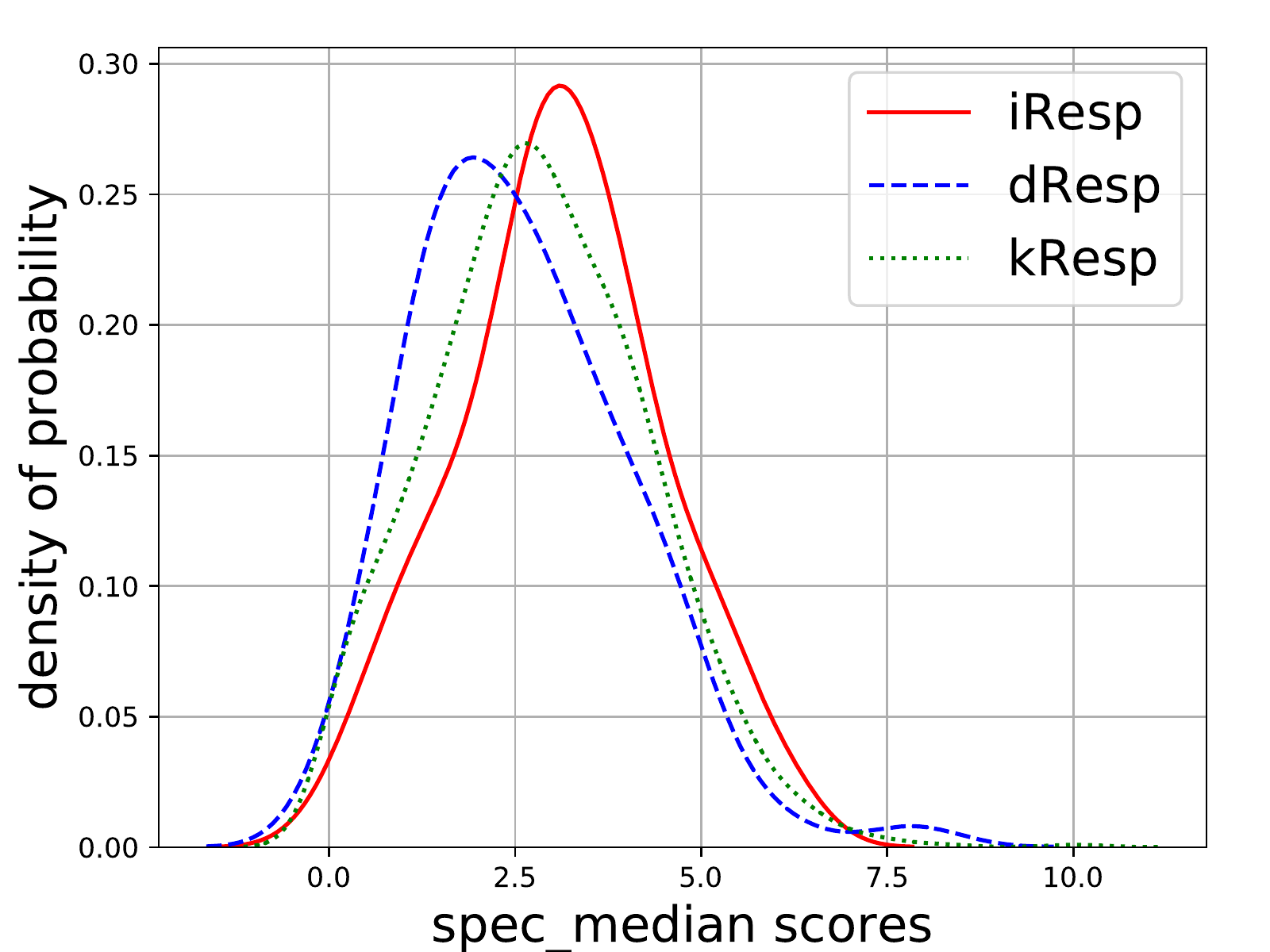}
    \caption{Smoothed distribution of specificity scores. }
    \label{fig:spec_distribution}
\end{figure}

We tokenize author responses with \texttt{nltk}, 
remove sentences with fewer than 10 tokens 
and rank the remaining sentences by their specificity scores.
All scores are normalized to $[0,10]$, with higher scores
meaning higher specificity.
The distribution of the specificity scores for author responses leading to increased, decreased and unchanged scores is illustrated in Fig. \ref{fig:spec_distribution}.
%\todo{IK: I changed here but cannot change the plot} 

\textbf{Top 10 sentences} are presented below\footnote{The examples are anonymized by replacing citations, venues, method names, exact scores, etc.\  with placeholders; we also include cases where our system has erroneously rated non-text data (i.e.\  tables).}, and they all receive a specificity score 10.
\textit{
\begin{itemize}
    \item 
   We have already checked it. We can change the sentence in the last paragraph of Section \#\#\# to ‘’Since the proposed method only substituted \#\#\# based on \#\#\#, then the naturalness of \#\#\# using the proposed method is better than \#\#\#. This method was used because we have to maintain the context; The result can be more than 100\% because we assume that the \#\#\# of original was 100\% while based on human judgement, there are possibility that the \#\#\# of resulting sentences using the proposed method is better than the original one.
    %We have already check about it; We can change the sentence in the last paragraph of 5.2.2 to ‘’Since the proposed method only substituted the word based on synonym word list, then the naturalness of paraphrased sentence using the proposed method is better than Gadag’s method”; This method was used because we have to maintain the context and for maintaining the context we have to find the same tag for a word in similar sentences; The result can be more than 100\% because we assume that the naturalness of original sentences was 100\% while based on human judgment, there are possibility that the naturalness of resulting sentences using the proposed method is better than the original one.
    \item  
    \#\#\#$|$ \#\#.\# $|$ \#\#.\# $|$   \#\#.\# $ |$   \#\#.\#  $|$ \#\#.\# $|$  \#\#.\#   $|$  \#\#.\# $|$    \#\#.\#   $|$
    %DisSent Books 5 $|$ 83.4 $|$ 81.8 $|$   93.4 $ |$   90.0  $|$ 82.5 $|$  87.0   $|$  0.8210 $|$    82.6   $|$
    \item  
    \#\#\#$|$ \#\#.\# $|$ \#\#.\# $|$   \#\#.\# $ |$   \#\#.\#  $|$ \#\#.\# $|$  \#\#.\#   $|$  \#\#.\# $|$    \#\#.\#   $|$
    %Our (RNN-CNN) $|$ 80.9 $|$ 82.2 $|$   94.1  $|$   88.8  $|$ 83.8 $|$  89.2   $|$  0.8735 $|$    84.7   $|$
    \item 
    There are two reasons why we mention that: (i) many papers exist, however, many previous papers made the same (or similar) conclusions, so some are picked up as representatives and (ii) because \#\#\# is a high-level conference, it's thought that there was no need to explain too much, and also because there are limited pages, space was wanted to be left to explain the analysis as detailed as possible and put focus on the analysis. 
    %There are two reasons why we wrote very few previous papers: (i) a lot of pervious papers exist, however, many previous papers made the same (or similar) conclusions, so some are picked up as representatives and (ii) because ACL is a high-level conference, it's thought that there was no need to explain too much, and also because there are limited pages, space was wanted to be left to explain the analysis as detailed as possible and put focus on the analysis. 
    \item 
    Other external knowledge sources apart from \#\#\# do not add much: In principle, all resources we used originate in \#\#\#, the difference is the degree of knowledge we use. The novelty in this work does not lie in the use of \#\#\# as a knowledge resource but more generally in the principled \#\#\# of the classes.
    %Other external knowledge sources apart from Wikipedia do not add much: In principle, all resources we used originate in Wikipedia, the difference is the degree of knowledge we use: from a list of entity names (i.e., Wikipedia title pages) to the named entities themselves (with their categories and set of possible names).The novelty in this work does not lie in the use of Wikipedia as a knowledge resource but more generally in the principled modularization of the knowledge classes.
    \item 
    We will include this discussion in the paper. Other \#\#\# models (e.g., \#\#\#; \#\#\#) can in theory predict \#\#\#, however, they are not directly applicable to \#\#\# since they cannot handle \#\#\# representations,  i.e., variables can refer to a \#\#\# representation (e.g., variable \#\#\# refers to an entire proposition and variable \#\#\# refers to a segment of meaning).
     %We will include this discussion in the paper· Other semantic parsing models (e.g., Dong and Lapata, 2016; Alvarez-Meils and Jaakkola, 2017) can  in theory to predict well-formed logical forms, however, they are not directly applicable to DRT parsing since they cannot handle scoped meaning   representations,  i.e., variables can refer to a block meaning representation (e.g., variable p1 refers to an entire  proposition and variable k1 refers to a segment of meaning).
    \item 
    As noted in our response to reviewer 3 - our results on the \#\#\# dataset of \#\#\# are on par with the \#\#\# model stated in the \#\#\# paper provided by reviewer 3 (which is a SOTA non-neural \#\#\# model) - although we used a very basic set of features and apply very limited task-specific tuning to our models.
    %As noted in our response to reviewer 3 - our results on the Spanish dataset of CoNLL2002 are on par with the Carreras et al 2002 model stated in the NAACL 2016 paper provided by reviewer 3 (which is a SOTA non-neural NER model) - although we used a very basic set of features and apply very limited task-specific tuning to our models.
    \item 
    Although the models used are general to all seq2seq generation problems, the heuristics we used to select \#\#\# are specific to generating the \#\#\# (take for example, the heuristic based on \#\#\# - it was motivated by the fact that \#\#\# have a higher readability, hence the network has to focus towards better readable information in the \#\#\# in order to generate \#\#\#).
    %Although the models used are general to all seq2seq generation problems, the heuristics we used to select the source sequence are specific to generating the news-blog title for a research paper (take for example, the heuristic based on readability - it was motivated by the fact that the blog titles have a higher readability, hence the network has to focus towards better readable information in the research article in order to generate the title for the news-blog post).
    \item 
    Because the size of the training data for \#\#\# task is very small, \#\#\# instances for \#\#\# task and \#\#\# instances for \#\#\# task, whereas the number of the parameters of the whole network is very big, we pre-training the \#\#\# network based on \#\#\#, released for \#\#\# task, and pre-training the \#\#\# network based on the training data for \#\#\# task.
    %Because the size of the training data for sentence-level QE task is very small, 23,000 instances for en-de task and 25,000 instances for de-en task, whereas the number of the parameters of the whole network is very big, we pre-training the encoder-decoder network based on bilingual parallel data, released for MT task, and pre-training the QE RNN network based on the training data for sentence-level QE task.
    \item 
    Related workshop and share tasks, including \#\#\# (collocated with \#\#\#), \#\#\# (collocated with \#\#\#), \#\#\# (collocated with \#\#\#), and \#\#\# (collocated with \#\#\#), show a great potential on applying NLP technologies to the \#\#\# domain.
    %Related workshop and share tasks, including SemEval 2017 Task 5 (collocated with ACL’17), FNP 2018 (collocated with LREC’18), FiQA 2018 (collocated with WWW’18), and ECONLP 2018 (collocated with ACL’18), show a great potential on applying NLP technologies to the financial domain.
\end{itemize}
}
        
\textbf{Bottom 10 sentences} are presented below. Their specificity scores
are all smaller than 0.001. 
%\todo{IK left as it is, it's quite anonymized already}
\textit{
\begin{itemize}
    \item 
     It would be a little difficult to build this connection.
    \item 
     It is not accurate and we will use 'obvious' instead.
    \item 
     We are not quite sure which part is not identical.
    \item 
     I will check that again and will write it as you said
    \item 
     Therefore we can see that they have no relation with each other.
    \item 
     We will try to do this in our future work.
    \item 
     But we do not see this as a weakness of our approach.
    \item 
     That is why we do not do that in the first submission.
    \item 
     So this is really true for all the "models".
    \item 
     Thank you very much for the reviews and for the very useful
\end{itemize}
}

\section*{Politeness Scores}
\begin{figure}
    \centering
    \includegraphics[width=0.45\textwidth]{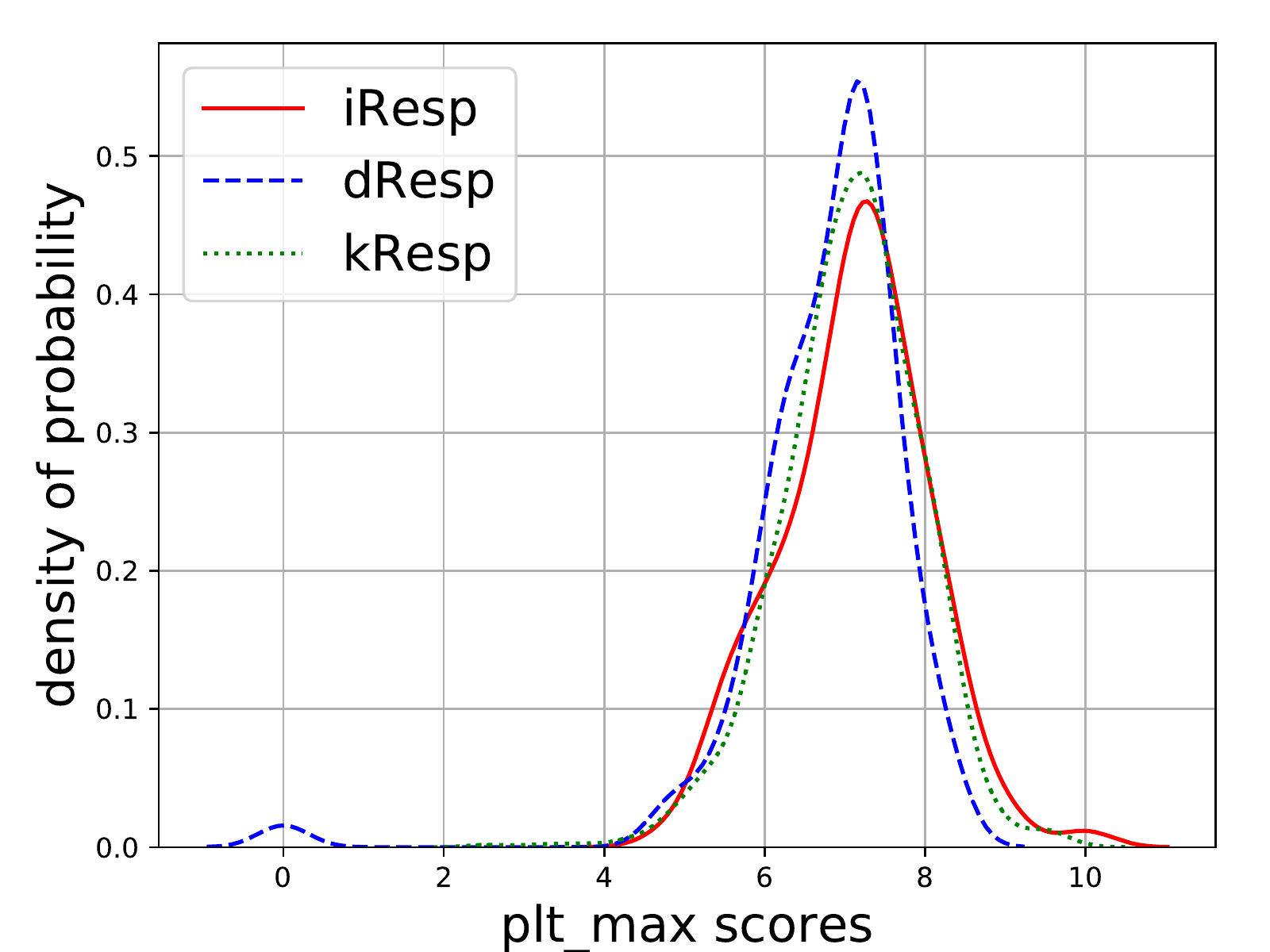}
    \caption{Smoothed distribution of politeness scores. }
    \label{fig:plt_distribution}
\end{figure}

We use the politeness scorer to rate the same
set of sentences as in the specificity evaluation.
We normalize all politeness scores to [0,10], with
higher values meaning higher politeness.
The distribution of the politeness scores is illustrated in Fig. 
\ref{fig:plt_distribution}.
\textbf{Top 10 sentences} and their politeness scores are presented below.

\textit{
\begin{itemize}
    \item (9.6) 
    We thank this reviewer for his helpful comments that help improving the paper.
    %We thank this reviewer for his helpful comments that help improving the paper.
    \item (9.5)
    Thanks for the suggestion, we found that in many cases the two sentences that are separated by \#\#\# also have similar patterns to \#\#\#, and the size of the dataset would be too small to train a representative \#\#\# model if we only picked out the separate sentences examples.
    %Thanks for the suggestion, we found that in many cases the two sentences that are separated by sentence-internal discourse markers also have similar patterns to the separate sentences examples, and the size of the dataset would be too small to train a representative DMP model if we only picked out the separate sentences examples.
    \item (9.5)
    Thank you for the helpful suggestion of including more qualitative results to more thoroughly understand the proposed approach.
    %Thank you for the helpful suggestion of including more qualitative results to more thoroughly understand the proposed approach.
    \item (9.4)
    We again thank the reviewer for the detailed and carefully constructed review and assure that the main concerns raised by the reviewer are fixable and we will fix them in the final version of the paper.
    %We again thank the reviewer for the detailed and carefully constructed review and assure that the main concerns raised by the reviewer are fixable and we will fix them in the final version of the paper.
    \item (9.4)
    Meanwhile, thanks for your suggestion for more in-depth discussion on \#\#\#
    %Meanwhile, thanks for your suggestion for more in-depth discussion on Wiseman2017.
    \item (9.3)
    We apologize for this error, and will correct this in the final version of the paper upon acceptance.
    %We apologize for this error, and will correct this in the final version of the paper upon acceptance.
    \item (9.3)
    An interesting alternative approach would be the one proposed by the reviewer, but we chose this model because we wanted to encourage the model to aggregate information from a variety of positions, and in our experience \#\#\# has trouble learning to \#\#\# in this way because by design \#\#\# focuses on one position only.
    %An interesting alternative approach would be the one proposed by the reviewer, but we chose this model because we wanted to encourage the model to aggregate information from a variety of positions (the lemma-pooling operation), and in our experience attention has trouble learning to pool in this way because by design (standard) attention focuses on one position only.
    \item 
     \#\#\#$|$ \#\#.\# $|$ \#\#.\# $|$   \#\#.\# $ |$   \#\#.\#  $|$ \#\#.\# $|$  \#\#.\#   $|$  \#\#.\# $|$    \#\#.\#   $|$
     %(0.84) DisSent Books 5 $|$ 83.4 $|$ 81.8 $|$   93.4 $ |$   90.0  $|$ 82.5 $|$  87.0   $|$  0.8210 $|$    82.6   $|$
    \item 
     \#\#\#$|$ \#\#.\# $|$ \#\#.\# $|$   \#\#.\# $ |$   \#\#.\#  $|$ \#\#.\# $|$  \#\#.\#   $|$  \#\#.\# $|$    \#\#.\#   $|$
    %(0.84) Our (RNN-CNN) $|$ 80.9 $|$ 82.2 $|$   94.1  $|$   88.8  $|$ 83.8 $|$  89.2   $|$  0.8735 $|$    84.7   $|$
    \item (9.2)
    Depends on the task and the characteristic of two datasets, each proposed method shows its effectiveness, e.g., the \#\#\# using the \#\#\# between two entities is appropriate for the \#\#\# task since \#\#\# is systematically organized.
    %Depends on the task and the characteristic of two datasets, each proposed method shows its effectiveness, e.g., the K-means-based approach using the embedding offset between two entities is appropriate for the automatic seed selection task since the taxonomy of relations in the part-whole dataset is systematically organized.
\end{itemize}
}

\textbf{Bottom 10} sentences and their politeness scores are presented below.
\textit{
\begin{itemize}
    \item (1.7)
    By comparing \#\#\# with \#\#\#-, we know whether employing a \#\#\# helps; By comparing \#\#\# with \#\#\#, we know whether employ a \#\#\# helps; By comparing \#\#\# with \#\#\#, we know whether the \#\#\# helps.
    %By comparing TreeGAN with TreeGAN-, we know whether employing a tree-structured classifier helps; By comparing TreeGAN with SeqGAN, we know whether employ a tree-structured, syntax-aware generator helps; By comparing TreeGAN with TreeGen, we know whether the adversarial training helps.
    \item (2.2)
    $\#\#\# = \#\#\# * \#\#\#$, where \#\#\# is a matrix of n samples with \#\#\# features followed by \#\#\# features, hence the size of $\#\#\#$ is $\#\#\#$.
    %$K_combined = [X1 X2] * [X1 X2]$, where [X1 X2] is a matrix of n samples with m1 HISK features followed by m2 BOSWE features, hence the size of $[X1 X2]$ is $n x (m1 + m2)$.
    \item (2.3)
    In other words, our coverage is \#\#\# times larger than theirs, so our proposed system can deal much better with the noise when learning \#\#\#.
    %In other words, our coverage is 99160/6000=16 times larger than theirs, so our proposed system can deal much better with the noise when learning multiple senses of infrequent words.
    \item (2.4)
    And another difference lies in the \#\#\# layer, which contains \#\#\#, so when we process \#\#\# in \#\#\# independently which encourages our model to learn diverse features.
    %And another difference lies in the word-embedding layer, which contains four parts, so when we process self-attention in four vector spaces independently which encourages our model to learn diverse features.
    \item (2.4)
    We will implement their method on our corpora and make some comparison with our method in the next version of our manuscript.
    %We will implement their method on our corpora and make some comparison with our method in the next version of our manuscript.
    \item (2.4)
    We are not giving up \#\#\# nor are we claimining that \#\#\# is more powerful.
    %We are not giving up continuous state representation nor are we claimining that discrete states are more powerful.
    \item (2.5)
    If our paper is accepted we will make sure additional relevant technical details are added.
    %If our paper is accepted we will make sure additional relevant technical details are added.
    \item (2.6)
     In response to your general remark: we can see how our discussion and conclusions would lead a reader to conclude that; rather, this paper is an exploration in an area that is, as you say, worth exploring.
    %argument 4:  In response to your general remark: we can see how our discussion and conclusions would lead a reader to conclude that; rather, this paper is an exploration in an area that is, as you say, worth exploring.
    \item (2.7)
    Our main contribution is introduction of \#\#\# without requiring neither supervision nor feature engineering.
    %Our main contribution is introduction of multiple latent relations without requiring neither supervision nor feature engineering.
    \item (2.7)
    The most salient problem encountered in our system is that a user might change \#\#\#, also brought up by R3 (Please refer to our response to weakness4 of R3).
    %The most salient problem encountered in our system is that a user might change her/his constraints, also brought up by R3(Please refer to our response to weakness4 of R3).
\end{itemize}
}

\section*{Convincingness Scores}

\begin{figure}
    \centering
    \includegraphics[width=0.45\textwidth]{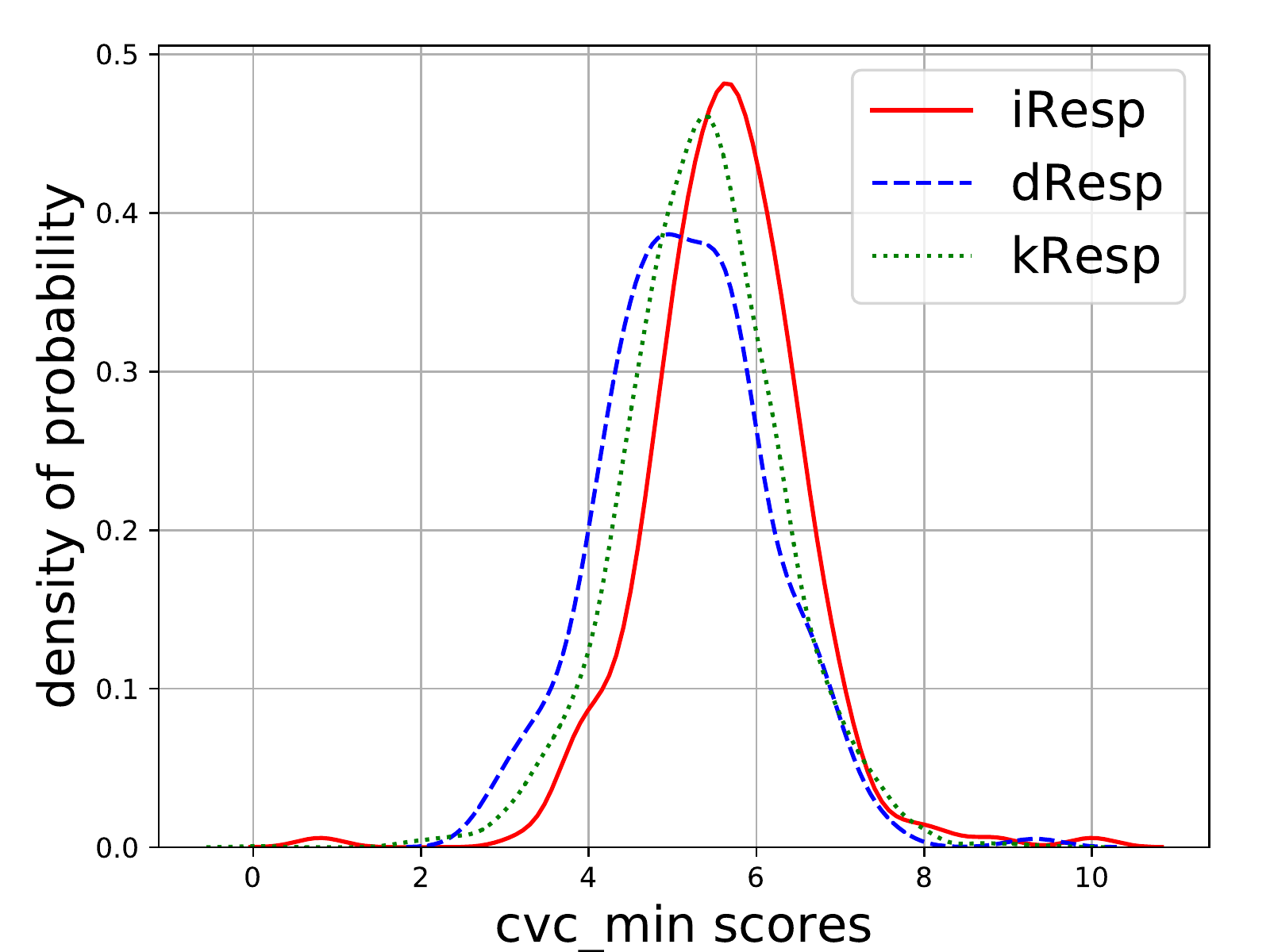}
    \caption{Smoothed distribution of the convincingess  scores.}
    \label{fig:cvc_distribution}
\end{figure}

We use the convincingness scorer to rate the same set of 
sentences as in the previous two studies.
The convincingness scores are normalized to $[0,10]$,
with higher values meaning higher convincingness.
The distribution of the convincingness scores is illustrated
in Fig.~\ref{fig:cvc_distribution}.
\textbf{Top 10 sentences} in terms of convincingness 
are presented below. 
All top 10 sentences' convincingness scores are above 9.8.

\textit{
\begin{itemize}
    \item 
    In the revision, we perform the evaluation of the model with \#\#\# and \#\#\#, respectively.
    %In the revision, we perform the evaluation of the model with POS tags and with time stamps, respectively.
    \item 
    Deepening the \#\#\# system would inevitably increase model parameters, and slow the training and decoding, which is not what we expect.
    %Deepening the NMT system would inevitably increase model parameters, and slow the training and decoding, which is not what we expect.
    \item 
    A technical document is defined as the document that assumes sufficient background knowledge and familiarity with the key technical or central/important terms in the document.
    %A technical document is defined as the document that assumes sufficient background knowledge and familiarity with the key technical or central/important terms in the document.
    \item 
    As reported in our paper, the success rate of our optimization algorithm is \#\#\# while, on average, only \#\#\#\% of words are altered.
    %As reported in our paper, the success rate of our optimization algorithm is 100\% while, on average, only 8.7% of words are altered.
    \item 
    The focus of this work is not a comparison of \#\#\# methods with \#\#\# methods, but how to mitigate the lack of labeled data problem in learning of a \#\#\# model.
     %The focus of this work is not a comparison of retrieval-based methods with generation-based methods, but how to mitigate the lack of labeled data problem in learning of a matching model.
    \item 
    Our model works well on datasets that are deemed small for deep architectures to work and belong to special domains for which \#\#\# is not possible.
    %Our model works well on datasets that are deemed small for deep architectures to work and belong to special domains for which transfer learning is not possible.
    \item %\todo{duplicate!}
    %The focus of this work is not a comparison of \#\#\# methods with \#\#\# methods, but how to mitigate the lack of labeled data problem in learning of a \#\#\# model.
    We conduct t-test and get the p value as \#\#\#, which shows good agreement.
   % The focus of this work is not a comparison of retrieval-based methods with generation-based methods, but how to mitigate the lack of labeled data problem in learning of a matching model.
    \item 
    Particularly, we will strive to improve the presentation quality and to make the draft more readable and better organized for more potential readers.
    %Particularly, we will strive to improve the presentation quality and to make the draft more readable and better organized for more potential readers.
    \item 
    Furthermore, \#\#\# can help \#\#\# to alleviate the performance degradation by \#\#\#.
    %Furthermore, KG-Net can help Corpus-Net to alleviate the performance degradation by knowledge distillation.
    \item 
    The \#\#\# experiments in Section \#\#\# show that our \#\#\# framework can achieve higher accuracy than the methods that rely on the same set of resources, while the state-of-the-art \#\#\# methods also require some other resources.
    %The BLI experiments in Section 4 show that our BLI framework can achieve higher accuracy than the methods that rely on the same set of resources, while the state-of-the-art BLI methods also require some other resources.
\end{itemize}
}

\textbf{Bottom 10 sentences} in terms of convincingness scores
are presented below\footnote{Note the large number of references to other responses and to the original reviews}.
Their convincingness scores are all below 0.01.
\textit{
\begin{itemize}
    \item
    "Weakness 3:""why ... report on ... the '\#\#\#' if you then dismiss it"""
   % "Weakness 3:""why ... report on ... the 'alignment-based dictionary' if you then dismiss it"""
    \item 
    It is **not** used in the **testing** (\#\#\#).
    \item 
    Annotator 1:  “Are you a citizen?” No  => Answer: No
    %Annotator 1:  “Are you a citizen?” No  => Answer: No
    \item 
    "Rev: ""It seems that ..."""
    %"Rev: ""It seems that ..."""
    \item 
    "Weakness 3:""how did you learn the embeddings? ...  \#\#\# model? How"
    %"Weakness 3:""how did you learn the embeddings? ...  Skip-gram model? How"
    \item 
    Please refer to the reply regarding Weakness argument 1 in Review 1.
    \item 
    “Are you over 21?” Yes => Answer: Yes
    \item 
    Please see our reply to Review 1’s weakness argument 3.
    \item 
    $[$Please see our response to R2’s argument 3$]$
    \item 
    We are sorry we didn’t explain the notation.
\end{itemize}
}
\label{sec:appendix}

\end{document}